\documentclass[10pt,twocolumn,letterpaper]{article}

\usepackage{cvpr}              

\usepackage{bm}

\usepackage{colortbl}
\usepackage[table]{xcolor}
\usepackage{booktabs}
\usepackage{multirow}
\usepackage[table]{xcolor} 

\usepackage{url}
\usepackage{graphicx} 
\usepackage{booktabs}
\usepackage{multirow} 
\usepackage{booktabs}
\usepackage[table]{xcolor}
\usepackage{makecell} 
\usepackage{pifont}
\usepackage{amsmath}
\usepackage{amssymb}
\usepackage{amsmath}
\usepackage{amssymb}
\usepackage{algpseudocode}
\usepackage{graphicx}
\usepackage{url}
\usepackage{bbm}
\usepackage{mathtools}
\usepackage{wrapfig}    
\usepackage{booktabs}   
\usepackage{multirow}   
\usepackage{array}      
\usepackage{graphicx}   
\usepackage{amssymb}    
\usepackage{capt-of} 
\usepackage{booktabs,multirow,makecell,xcolor,tabularx}

\usepackage[ruled,vlined]{algorithm2e}
\usepackage{enumitem}

\DeclareMathOperator{\hit}{Hit}
\newcommand{\Hit}{\hit_{3\times3}}  %
\newcommand{\vecb}[1]{\bm{#1}}            
\newcommand{\matb}[1]{\bm{#1}}            
\newcommand{\Mgt}{\matb{M}_{\texttt{gt}}} 

\DeclareMathOperator{\clip}{clip}
\definecolor{mycolor}{RGB}{0, 128, 128}
\definecolor{mycolor3}{RGB}{129, 216, 207}
\definecolor{mycolor2}{RGB}{203, 109, 81}
\usepackage{bm}
\usepackage{booktabs, multirow, makecell, xcolor, pifont, caption}
\usepackage{arydshln}
\usepackage{bm}

\definecolor{cvprblue}{rgb}{0.21,0.49,0.74}
\usepackage[pagebackref,breaklinks,colorlinks,allcolors=cvprblue]{hyperref}

\newcommand{\best}[1]{\textbf{#1}}

\newcommand{\offset}[2]{\texttt{<OFF$_{#1,#2}$>}}
\newcommand{\del}{\texttt{<DELETE>}}
\usepackage[accsupp]{axessibility}
\def\paperID{6454} 
\def\confName{CVPR}
\def\confYear{2026}

\title{Grounding Everything in Tokens for Multimodal Large Language Models}

\vspace{-2em}
\author{
    Xiangxuan Ren$^{1}$ \quad Zhongdao Wang$^{2}$ \quad Liping Hou$^{2}$ \quad Pin Tang$^{1}$ 
    \quad Guoqing Wang$^{1}$ \quad Chao Ma$^{1}$\footnotemark[2]\\ 
    ${}^{1}$ MoE Key Lab of Artificial Intelligence, AI Institute, Shanghai Jiao Tong University \\
    ${}^{2}$ Central Research Institute, Huawei\\
    {\tt\small \{bunny\_renxiangxuan, pin.tang, guoqing.wang, chaoma\}@sjtu.edu.cn} \\
    {\tt\small zhongdwang@gmail.com, houliping1@huawei.com}
    }

\begin{document}
\twocolumn[{
\renewcommand\twocolumn[1][]{#1}
\maketitle
\vspace{-3em}
\begin{center}
    \captionsetup{type=figure}
    \includegraphics[width=0.94\textwidth]{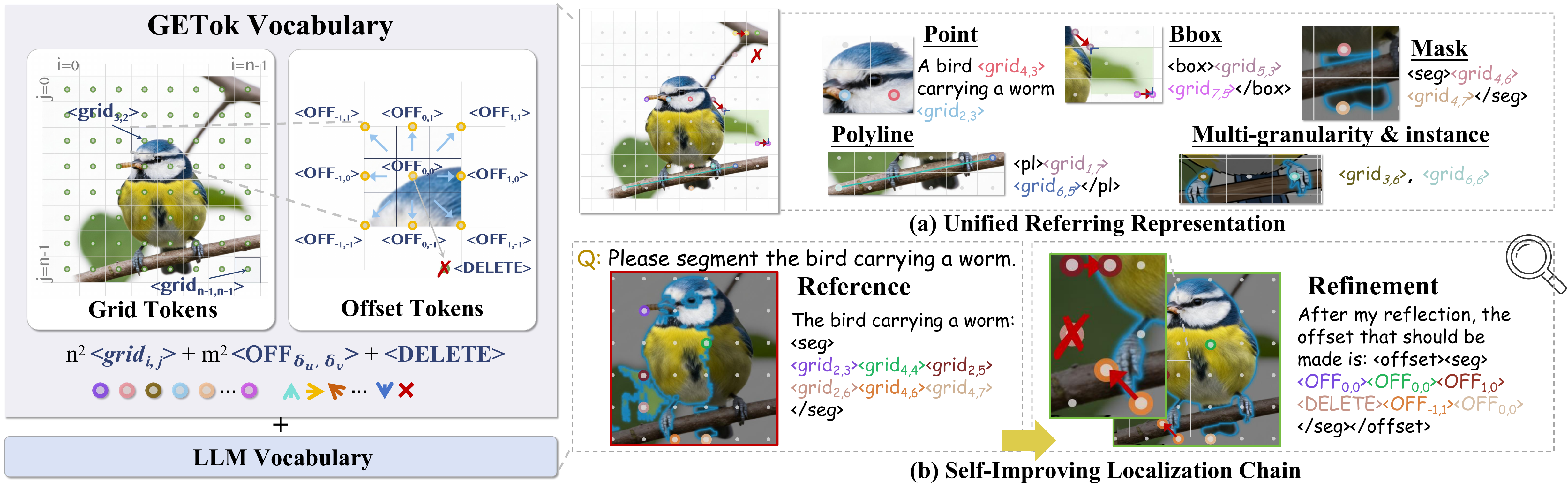}
    \captionof{figure}{
    Overview of GETok. GETok equips MLLMs with pre-defined, learnable discrete tokens tied to uniformly distributed anchor points on the image plane, enabling unified grounding from diverse inputs such as text, points, bounding boxes, and segmentation masks. 
    A localization refinement scheme further supports coarse-to-fine correction and iterative recovery from initial grounding errors.
    }
    \label{fig:teaser}
\end{center}
}]

\begingroup
\renewcommand{\thefootnote}{\textdagger}
\footnotetext{\scriptsize Corresponding author.}
\endgroup

\begin{abstract}

Multimodal large language models (MLLMs) have made significant advancements in vision understanding and reasoning. However, the autoregressive Transformer architecture used by MLLMs requires tokenization on input images, which limits their ability to accurately ground objects within the 2D image space. This raises an important question: how can sequential language tokens be improved to better ground objects in 2D spatial space for MLLMs?
To address this, we present a spatial representation method for grounding objects, namely GETok, that integrates a specialized vocabulary of learnable tokens into MLLMs. GETok first uses grid tokens to partition the image plane into structured spatial anchors, and then exploits offset tokens to enable precise and iterative refinement of localization predictions. By embedding spatial relationships directly into tokens, GETok significantly advances MLLMs in native 2D space reasoning without modifying the autoregressive architecture.
Extensive experiments demonstrate that GETok achieves superior performance over the state-of-the-art methods across various referring tasks in both supervised fine-tuning and reinforcement learning settings. Project page: \url{https://getokpage.github.io}
\end{abstract}  
\vspace{4pt}
\section{Introduction}
Recent years have witnessed significant advancements in multimodal large language models (MLLMs)~\citep{alayrac2022flamingo,li2022blip,liuVisualInstructionTuning2023,liBLIP2BootstrappingLanguageImage2023,gpt4v,VLM:Gemini,zhu2025internvl3,liu2024llava} concerning vision understanding, reasoning, and interaction. The impressive successes of autoregressive Transformers in language modeling~\cite{openaiGPT4TechnicalReport2023,devlin2018bert,brown2020language,gao2023llama} have established them as the foundational architecture for MLLMs. Autoregressive Transformers typically require tokenization of input images, similar to that used for text. However, this image tokenization often leads to a substantial loss of spatial information~\cite{jiang2024chatrex,ma2024clawmachine,sun2023emu}. As a result, current MLLMs encounter a notable limitation in their ability to reason accurately about precise spatial localization.

\begin{figure}[t]
    \centering
    \includegraphics[width=\linewidth]{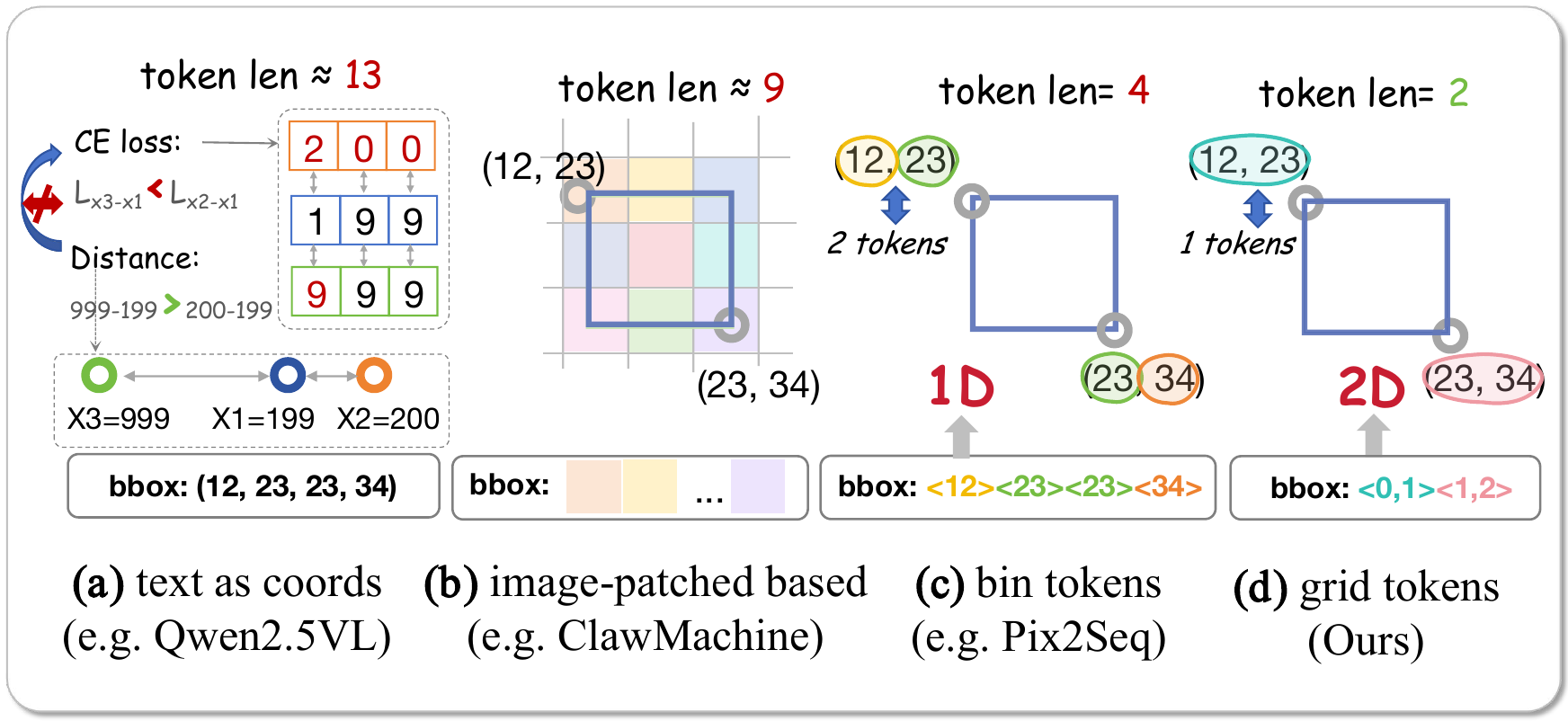}
    \vspace{-1.5em}
    \caption{
    Comparison of token-based representations for grounding objects in MLLMs. Note that 2D grid tokens preserve spatial topology with shorter sequences than coordinate-, patch-, or 1D bin-based formulations.
    }
    \label{fig:Intro}
    \vspace{-1.5em}
\end{figure}

Numerous efforts have been made to address this issue. 
One straightforward solution is to use text to describe object locations~\citep{chen2023shikra,you2023ferret,internvl},  as demonstrated by Qwen-VL~\cite{bai2025qwen2}. However, this text-based approach struggles to preserve spatial topology~\cite{su2025patch}, resulting in large syntactic overhead and tokenization bias, as illustrated in Fig.~\ref{fig:Intro}(a). 
More recently, a number of methods directly project image patches into visual tokens via linear projection~\cite{ma2024clawmachine,jin2023unified,sun2023emu,emu3}, as shown in Fig.~\ref{fig:Intro}(b). 
However, the patch size is fixed by the image encoder, and the linear projection is tied to this patch partition, entangling texture with geometry and often confusing texture-similar objects at different spatial locations.
Alternatively, bin-based methods~\cite{chen2021pix2seq, wu2024visionllm,wang2022ofa} use one-dimensional bins to describe bounding boxes for grounding objects (Fig.~\ref{fig:Intro}(c)). While promising, slight changes in one-dimensional indices do not accurately reflect smooth changes in 2D topology, so bin tokens benefit less from recent reinforcement learning schemes such as GRPO~\cite{shao2024deepseekmath}, where small action changes can unexpectedly cause large reward fluctuations.

In this work, we identify the core challenge in advancing MLLMs toward precise 2D reasoning as establishing a reliable mapping between discrete sequential tokens and continuous 2D space.
As such, we propose \emph{GETok} for \emph{G}rounding \emph{E}very object in \emph{Tok}ens in MLLMs via a set of learnable \textit{spatial vocabulary} terms. As shown in the left panel of Fig.~\ref{fig:teaser},  our GETok comprises two core types of tokens:
i) \textbf{Grid tokens} first establish a structured spatial topology by discretizing the image plane into an \(n\times n\) uniform grid. Each grid cell is associated with a learnable token added to the model's vocabulary, yielding a set of \emph{spatial anchors}, each of which is responsible for referring to objects within its local region. 
While this 2D lattice provides native spatial awareness, it introduces a vocabulary bottleneck, as the number of tokens grows quadratically with increasing resolution. 
ii)~\textbf{Offset tokens} overcome this issue and refine spatial reasoning via a set of discrete displacement vectors together with a \del \ token. 
Building upon the structural regularity of grid tokens, offset tokens enable high-precision spatial refinement at a minimal vocabulary cost. For example, a \(32^2\) anchor grid can be upgraded to \(64^2\) effective precision by adding ten offset tokens, instead of introducing \(64^2 - 32^2 = 3072\) new grid tokens. 
Furthermore, the use of offset tokens yields an emergent benefit of progressive localization refinement. 
Because the \texttt{<DELETE>} token can recursively reject errors during localization, it transforms the process from a one-shot prediction into an iterative reasoning approach.

GETok presents three significant advantages over state-of-the-art methods:
\emph{First}, GETok provides a unified representation for various tasks, ranging from points to masks, all within a standard autoregressive framework (Fig.~\ref{fig:teaser}(a)). 
This integration eliminates the need for task-specific modules, simplifying the architecture while ensuring generalizability and precision.
\emph{Second}, the integrated offset mechanism facilitates self-correction through iterative refinement. This feature allows the model to adjust its spatial predictions, addressing a common limitation in existing methods where initial errors often go uncorrected (Fig.~\ref{fig:teaser}(b)).
\emph{Third}, the geometric foundation that correlates token shifts with smooth spatial changes creates a low-entropy action space. This results in stable reward landscapes and more efficient exploration, significantly enhancing policy optimization compared to unstructured representations.
Building on these advantages, we introduce a novel \emph{self-improving reinforcement learning framework} that explicitly models spatial dynamics and employs GRPO-style preference optimization to refine locations through iterative self-correction.
Comprehensive experiments across various referring benchmarks show that our GETok achieves superior performance under both supervised fine-tuning and reinforcement learning settings.

In summary, the main contributions of this work are:
\begin{itemize}[leftmargin=1.2em,labelsep=0.5em,itemsep=0.2em,topsep=0.2em]

\item We propose a lexical spatial representation that embeds a vocabulary of learnable tokens to empower MLLMs to accurately reason over native 2D spatial space without modifying autoregressive frameworks.
\item We develop a localization refinement scheme on top of offset tokens that provides coarse-to-fine correction and iterative recovery from initial grounding errors.
\item We propose a geometry-aware policy optimization framework to perform self-improving reinforcement learning to facilitate spatial reasoning. 
\end{itemize}

\begin{figure*}[t]
  \centering
  \includegraphics[width=\linewidth]{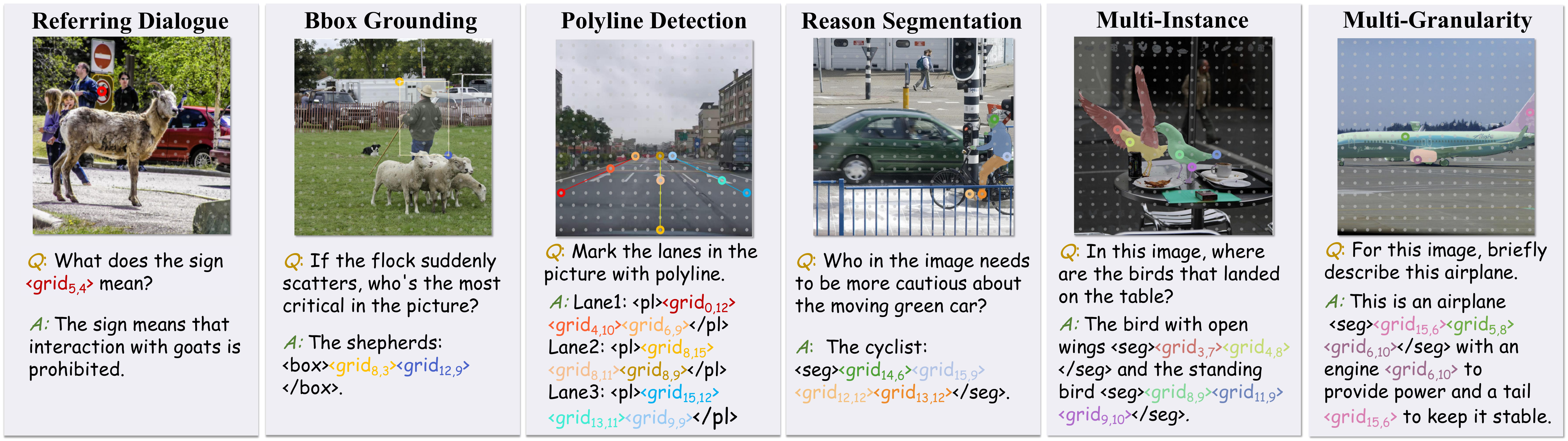}
  \caption{GETok supports both input and output references with multiple format conversions, including boxes, polylines, and masks. It is seamlessly compatible with multi-instance and multi-granularity capabilities. Best viewed in color.
  }
  \label{fig:unified_format}
  \vspace{-1em}
\end{figure*}

\section{Related Work}
\paragraph{MLLMs for Visual Grounding.}
Enabling MLLMs to understand, manipulate, and output image regions is a central goal for vision-language intelligence~\cite{liu2020consnet,liu2024catnet,r2tuning,liu2022umt}.
Current methods primarily follow two training paradigms: 
Under the SFT paradigm, methods have explored diverse referring representations, including points~\cite{deitke2025molmo}, bounding boxes~\cite{chen2023shikra}, and masks~\cite{lisa}.
Early approaches demonstrated that text-based coordinate representations~\cite{chen2023shikra, you2023ferret, wang2023cogvlm, zhan2024griffon, jiang2024chatrex, ma2024groma, jiang2025referringperson, zhu2025internvl3, guo2025seed1} can enable basic referring dialogue, but these methods suffered from ambiguous region-text alignment. This limitation prompted methods like GPT4RoI~\cite{zhang2023gpt4roi} to introduce specialized region modules~\cite{chenPositionEnhancedVisualInstruction2023,ma2024groma,tian2024chatterbox} for improved alignment at the cost of architectural complexity. 
For complex shape representation, mask-based referring has emerged as a promising direction. Ferret~\cite{you2023ferret} and Osprey~\cite{yuan2024osprey} designed specialized pooling mechanisms for irregular mask inputs, while LISA~\cite{lisa} pioneered an embedding-as-mask paradigm using dedicated segmentation tokens that trigger external decoders, a design that has inspired a series of subsequent works~\cite{gsva,glamm,unipixel2025}. 
Nevertheless, SFT-based approaches remain limited when handling complex spatial reasoning tasks. The emergence of RL-based methods has demonstrated significant potential for overcoming these limitations~\cite{acuna2025long,openai2024o1,you2025seg}, such as VLM-R1~\citep{shen2025vlmr1} and Visual-RFT~\citep{liu2025visual}. Approaches like Seg-Zero~\citep{liu2025segzero} and VisionReasoner~\citep{liu2025visionreasoner} further demonstrate the effectiveness of decoupled architectures where reasoning chains generate prompts for external segmenters, establishing RL as a promising direction for mask output.
In this work, we revisit both paradigms through the lens of a lexical spatial vocabulary and a self-improving RL framework to endow MLLMs with native 2D spatial reasoning.

\vspace{-5mm}
\paragraph{Token-based Referring Representation.} 
A complementary research direction focuses on unifying visual and linguistic representations through discrete tokenization.
One line of work employs special tokens to aggregate positional information~\cite{lisa,you2023ferret,gsva,groundhog,zhang2023nextchat,zhang2024ferret}, which encode image regions through dedicated tokens but typically require additional architectural modifications to process this spatial information.
Another line leverages image patches as visual tokens~\cite{sun2023emu,sun2023generative,jin2023unified,team2024chameleon,sun2024autoregressive,ma2024clawmachine}, though this remains tightly coupled with the visual encoder~\cite{radford2021clip}, limiting transferability.
Bin-based methods~\cite{chen2021pix2seq,wu2024visionllm,yang2022unitab} discretize coordinates into tokens selected from a fixed vocabulary. While these methods introduce learnable spatial tokens, they remain limited to 1D indexing, which does not adequately capture explicit 2D structural relationships. \emph{The closest related work, Kosmos-2~\cite{peng2023kosmos}}, also envisions 2D spatial tokens but is restricted to basic bounding box grounding, lacking the ability to support more complex spatial grounding tasks.
In contrast, our method aims to establish a general-purpose spatial lexicon in which grid tokens create an explicit 2D lattice. This lattice enables referencing across various formats, including points, boxes, polylines, and masks (See Fig.~\ref{fig:unified_format}). Moreover, offset tokens facilitate coarse-to-fine corrections with minimal vocabulary expansion. This effectively improves localization precision while preserving architectural simplicity.

\section{GETok: Grounding Everything in Tokens}
\subsection{Overview}

\begin{figure}[t]
  \centering
  \includegraphics[width=0.96\linewidth]{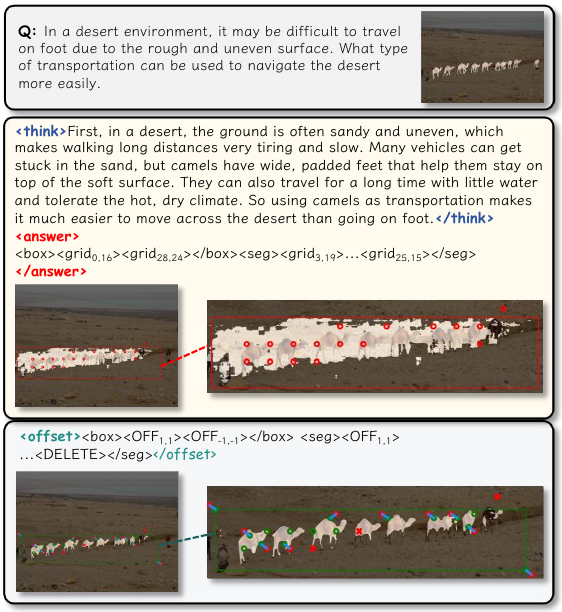}
  \caption{Example of the propose-and-refine mechanism in GETok. Grid tokens provide coarse localization, while offset tokens enable precise adjustment.}
  \vspace{-1em}
  \label{fig:GETok_case}
\end{figure}
\begin{figure*}[t]
    \centering
    \includegraphics[width=0.95\linewidth]{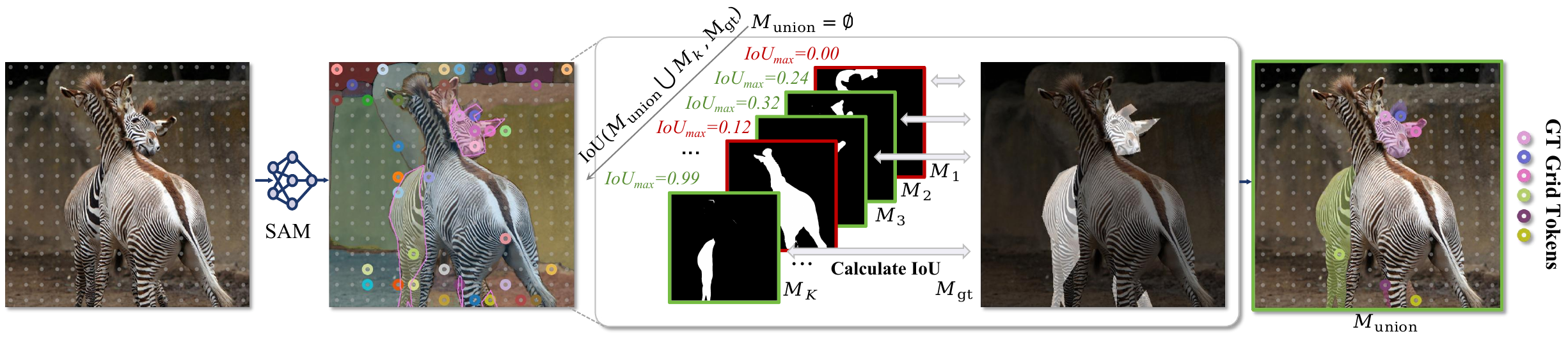}
    \vspace{-2mm}
    \caption{{\textbf{We use a greedy algorithm to generate the ground-truth grid tokens referring to the ground-truth mask.} This conversion automatically transforms continuous masks into discrete tokens, enabling scalable data expansion.} 
    }
    \label{fig:dataset_pipeline}
   \vspace{-1em}
\end{figure*}

To endow MLLMs with the ability to interpret and generate spatial references in a native token-based manner, we propose to augment their vocabulary with a set of learnable spatial tokens. Specifically, we first use grid tokens to discretize the image into $n\times n$ anchors:
$\mathcal{T}_{\text{grid}} = \{\texttt{<grid}_{i,j}\texttt{>} \mid i,j \in \{0,\dots,n-1\}\}$. Then, we use offset tokens to refine local positions: 
$\mathcal{T}_{\text{offset}} = \{\texttt{<OFF}_{\delta_u,\delta_v}\texttt{>}\} \cup \{\texttt{<DELETE>}\}$, where $\delta_u,\delta_v \in \{-1,0,1\}$.
The complete vocabulary $\mathcal{V} = \mathcal{V}_{\mathrm{LLM}} \cup \mathcal{T}_{\text{grid}} \cup \mathcal{T}_{\text{offset}}$ facilitates spatial reasoning as precise spatial pronouns. These two types of tokens collectively reason about localization through a \emph{propose-and-refine} chain. Fig.~\ref{fig:GETok_case} illustrates an example where a complex mask cannot be adequately represented by bounding boxes. When using grid tokens to initiate coarse proposals and offset tokens to refine them iteratively, we succeed in constructing precise representations for intricate masks. 
We next describe how GETok is instantiated under supervised fine-tuning and reinforcement learning settings.

\begin{figure*}
    \centering
    \includegraphics[width=0.96\linewidth]{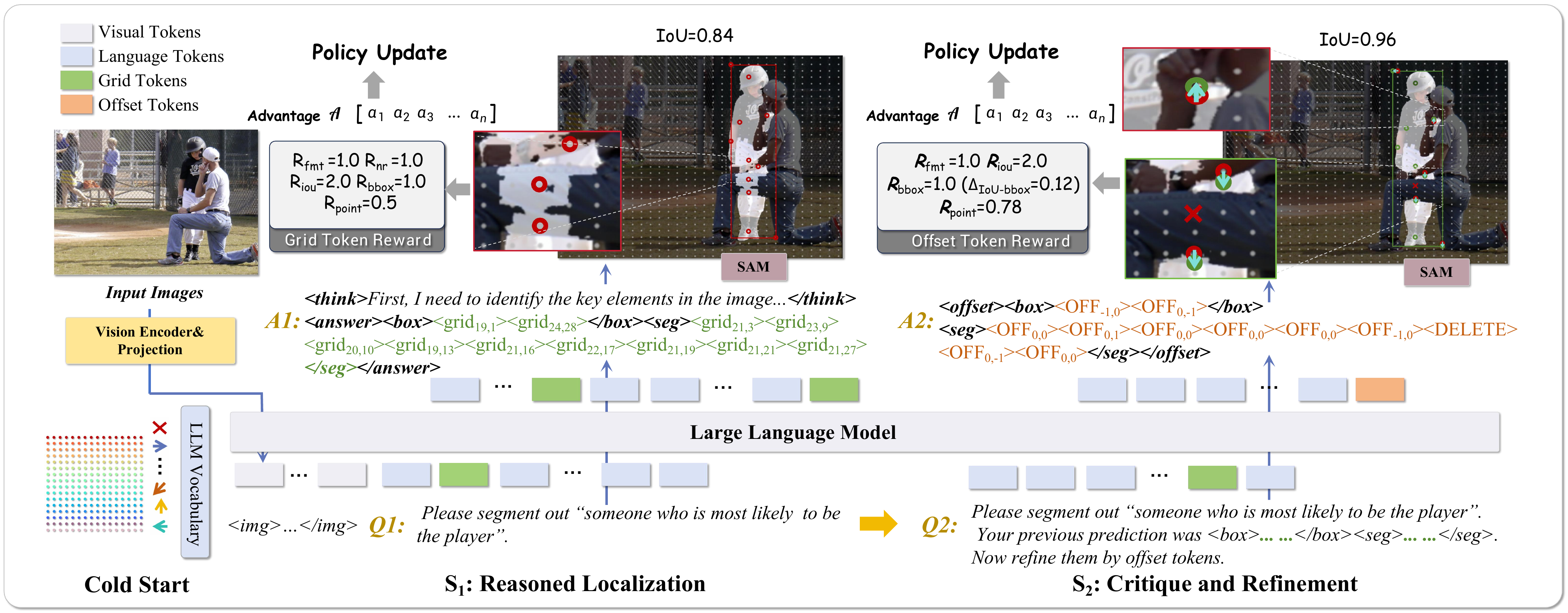}
    \caption{\textbf{Overview of the Self-Improving RL Framework.} Our framework models 2D spatial localization as a two-step generative task. First, grid tokens are generated to propose anchor regions in the image. Second, offset tokens refine the region proposals to precise points. }
    \label{fig:arch}
    \vspace{-10pt}
\end{figure*}
\subsection{Supervised Fine-Tuning}
\label{sec:sft}
Since GETok does not require modifying the architecture of base MLLMs, the key to applying supervised fine-tuning (SFT) lies in constructing training data. As such, SFT can effectively leverage the GETok vocabulary through automated annotation conversion and sequence simulation. While grid tokens provide a unified representation for points, bounding boxes, and polylines through straightforward mappings, as shown in Fig.~\ref{fig:unified_format}, two core challenges remain: (1) how to construct discrete point representations for dense masks, and (2) how to create effective training data for offset tokens to perform localization refinement.

\subsubsection{Greedy Mask-to-Token Conversion}
\label{sec:seg}
When converting dense masks into discrete points, current token-based approaches generally use single points, bounding boxes, combinations of bounding boxes with one or two fixed points, or randomly sampled points within a mask~\cite{deitke2025molmo,liu2025visionreasoner,liu2025segzero}. However, we observe that these formats often exhibit significant redundancy and ambiguity, especially when dealing with multiply-connected mask regions. 
To address this issue, we develop a greedy algorithm to facilitate the transformation from masks to grid tokens. Importantly, this conversion process is training-free and does not incur any additional computational costs or require changes to the model architecture.

As illustrated in Fig.~\ref{fig:dataset_pipeline}, we initially input the image along with \( n^2 \) grid points as prompts into the SAM~\cite{sam}. This process generates \( K \) masks, denoted as \( \mathcal{M} = \{\mathbf{M}_1, \ldots, \mathbf{M}_K\} \). Each mask corresponds uniquely to an input grid, defined by a mapping \(\theta: \{i\}_{i=1}^{n^2} \rightarrow \{k\}_{k=1}^{K}\). Typically, \( K < n^2 \) because of mask deduplication during post-processing.
Given a ground-truth mask \( \mathbf{M}_{\texttt{gt}} \), our goal is to identify \emph{a minimal set of} grid points such that the union of their corresponding masks approximates \( \mathbf{M}_{\texttt{gt}} \). Formally, this objective is written as:
\begin{equation}
\label{eq:mask-match}
\begin{split}
&\boldsymbol{\pi}^\star 
= \arg\min_{\boldsymbol{\pi}\in\{0,1\}^{n^2}}\|\boldsymbol{\pi}\|_{0} \\
&\text{s.t.} \quad \operatorname{IoU}\!\left(\mathbf{M}_{\mathrm{gt}},\,\bigcup_{k:\,\pi_k=1} \mathbf{M}_{\theta(k)}\right)\ge \tau .
\end{split}
\end{equation}
Here, \( \boldsymbol{\pi} \) is a binary selection vector over the grid tokens, and \( \tau \) is a quality threshold that ensures a minimum Intersection-over-Union (IoU).
Eq.~\ref{eq:mask-match} defines a constrained multi-objective optimization problem. To solve it efficiently, we develop a simple yet effective greedy algorithm starting with \( \boldsymbol{\pi} = \mathbf{0} \), \( \mathbf{M}_{\texttt{union}} = \mathbf{0} \), and \( \text{IoU}_\text{max} = 0 \). 
First, we compute the IoUs between \( \mathbf{M}_{\texttt{gt}} \) and all \( K \) mask proposals, sorting them in descending order. Then, we iterate through all masks. For the \( k \)-th iteration, we calculate \( \text{IoU}^{*} = \text{IoU}(\mathbf{M}_{\texttt{union}} \cup \mathbf{M}_k, \mathbf{M}_{\texttt{gt}}) \). If \( \text{IoU}^{*} > \text{IoU}_{\text{max}} \), we update \( \pi_k \leftarrow 1 \), \( \mathbf{M}_{\texttt{union}} \leftarrow \mathbf{M}_{\texttt{union}} \cup \mathbf{M}_k \), and set \( \text{IoU}_{\text{max}} \leftarrow \text{IoU}^{*} \). 
At the end of the iterative process, we obtain an approximately optimal \( \boldsymbol{\pi}^{*} \) that identifies the grid points corresponding to the ground truth mask. Finally, a mask can be represented as an unordered sequence of matched grid tokens, for example, \texttt{<seg><grid$_{{i_1},{j_1}}$>...<grid$_{{i_n},{j_n}}$></seg>}.

\subsubsection{Offset-Aware Dataset Construction}
\label{sec:offset_dataset}
To generate high-quality training data for offset tokens, we develop a systematic approach that categorizes grid points based on their spatial relationships to mask boundaries. 
Using morphological operations scaled to the offset step size, we define four distinct regions around each mask boundary:
i) \textbf{Inside}: Stable interior points mapped to zero offset (\offset{0}{0});
ii) \textbf{Ring}: Boundary-proximal exterior points requiring non-zero offsets;
iii) \textbf{Far}: Distant negatives mapped to deletion (\del);
iv) \textbf{Hard-Delete}: Challenging edge cases also mapped to \del.
Each grid point is assigned to exactly one region through an ordered decision rule that prioritizes high-value training cases. Training pairs are sampled with a bias toward \textsc{Inside} and \textsc{Ring} regions where offset corrections provide the most learning value.

This procedure yields a variable number \(K\) of grid–offset token pairs per image for supervised training. 
Empirically, this simulated supervision outperforms real-generated alternatives by focusing on boundary-proximal scenarios, creating a curated set of high-value training cases that foster effective refinement strategies. Detailed algorithms are provided in the supplementary.

\begin{table*}[t]
\centering
\setlength{\tabcolsep}{9pt} 
\renewcommand{\arraystretch}{0.95}
\caption{\textbf{Referring Expression Segmentation} (RES) results on the ReasonSeg and RefCOCO (+/g) datasets.}
\label{tab:res}
\resizebox{0.96\linewidth}{!}{ 
\begin{tabular}{lccccccccccccccccc}
\toprule
\multirow{2}{*}{\textbf{Methods}} & \multirow{2}{*}{\makecell{\textbf{Training }\\\textbf{Mask Dec.}}}&& \multicolumn{2}{c}{\textbf{ReasonSeg}}&& \multicolumn{3}{c}{\textbf{refCOCO}} && \multicolumn{3}{c}{\textbf{refCOCO+}} && \multicolumn{2}{c}{\textbf{refCOCOg}} && \multirow{2}{*}{\textbf{Avg.}}\\

\cmidrule(lr){4-5}\cmidrule(lr){7-9} \cmidrule(lr){11-13} \cmidrule(lr){15-16}

& &&\textbf{Val.} &\textbf{Test} && \textbf{Val.} & \textbf{T-A} & \textbf{T-B} && \textbf{Val.} & \textbf{T-A} &\textbf{T-B} && \textbf{Val.} & \textbf{Test} && \\
\midrule
\midrule
\multicolumn{18}{c}{\textit{—— Supervised Fine-Tuning Models ——}}\\
LAVT \cite{yang2022lavt} &\ding{52} &&- &- &&72.7 & 75.8 & 68.8 && 62.1 & 68.4 & 55.1 && 61.2 & 62.1 && - \\
ReLA \cite{liu2023gres} &\ding{52} &&- &- &&73.8 & 76.5 & 70.2 && 66.0 & 71.0 & 57.7 && 65.0 & 66.0 && - \\
CRIS \cite{wang2022cris} &\ding{52} &&- &- &&70.5 & 73.2 & 66.1 && 65.3 & 68.1 & 53.7 && 59.9 & 60.4 && -\\
PixelLM \cite{pixellm} &\ding{52}&&- &- &&73.0 & 76.5 & 68.2 && 66.3 & 71.7 & 58.3 && 69.3 & 70.5 && - \\
LISA \cite{lisa} &\ding{52} &&44.4 &36.8 &&{76.0} & {78.8} & {72.9} && 65.0 & 70.2 & 58.1 && \best{69.5} & 70.5 && 64.2 \\
Qwen2.5-VL-7B \cite{bai2025qwen2} &\ding{56} &&55.4 &51.5 &&{72.5} & 76.4 & 70.0 && 64.3 & 70.5 & 58.4 && 68.1 & 69.9 &&65.7 \\
\rowcolor{violet!6}{GETok-SFT-grid} &\ding{56}&&{58.1} &{54.4} && {74.3} & {77.9} & {72.3} && {65.6} & {71.9} & {58.8} &&68.0 &\best{70.9} && {67.2} \\
\rowcolor{violet!6}{GETok-SFT} &\ding{56}&&\best{59.2} &\best{55.8} && \best{76.1} & \best{79.2} & \best{73.2} && \best{66.4} & \best{72.3} & \best{59.9} &&69.4 &\best{70.9} && \best{68.2} \\
\midrule
\multicolumn{18}{c}{\textit{—— Reinforcement Learning Models ——}}\\
Seg-Zero \cite{liu2025segzero} &\ding{56} &&62.6 &57.5 &&- & 80.3 & - && - & 76.2 & - && - & 72.6 && 69.8 \\
SAM-R1 \cite{samr1} &\ding{56} &&64.0 &60.2 &&- & 79.2 & - && - & 74.7 & - && - & 73.1 && 70.2 \\
VisionReasoner \cite{liu2025visionreasoner} &\ding{56} && \best{66.3} &63.6 &&- & 79.3 & - && - & 72.2 & - && - & 72.2 && 70.7 \\

\rowcolor{violet!6}{GETok-R1-grid} &\ding{56}&& 64.2 & 63.7 && {-} & {79.8} & {-} && {-} & {74.3} & {-} &&- &73.9  && {71.2} \\
\rowcolor{violet!6}{GETok-R1} &\ding{56}&& {65.9} & \best{64.2} && {-} & \best{80.8} & {-} && {-} & \best{77.4} & {-} &&-  &\best{75.2}  && \best{72.7} \\
\bottomrule
\end{tabular}
}
\end{table*}

\begin{figure*}[t]
    \centering
    \includegraphics[width=0.95\linewidth]{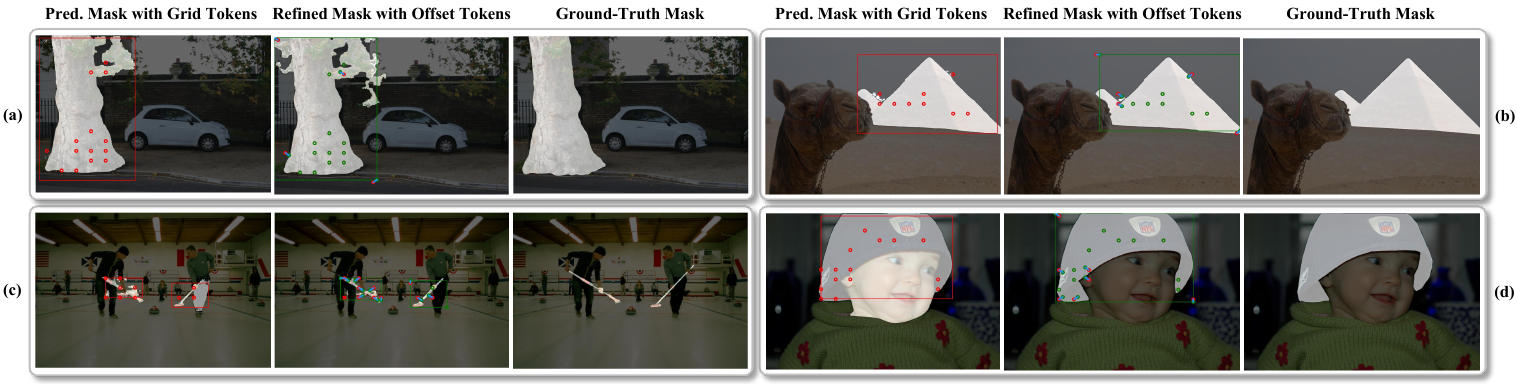}
    \caption{\textbf{GETok Qualitative Results on RES~\cite{yu2016refcoco}.} We visualize the two-step localization process: red dots are grid-token proposals, blue lines show the applied offset vectors, and green dots represent the final offset-refined points. Our method demonstrates adaptive corrections, achieving precise localization across diverse scenarios, including small objects and complex shapes.}
    \label{fig:visual1}
    \vspace{-1em}
\end{figure*}
\subsection{Self-Improving Reinforcement Learning}
The structured nature of GETok offers an ideal framework for reinforcement learning due to its 2D lattice organization, which creates a geometrically grounded action space rich in spatial semantics. We introduce a novel self-improving reinforcement learning framework that utilizes the grid-offset hierarchy of GETok to enable iterative self-correction. Unlike traditional RL methods that optimize a single-shot prediction, our method incorporates a multi-turn refinement process, allowing the model to critique and adjust its spatial predictions using offset tokens and a \texttt{<DELETE>} command.

As illustrated in Fig.~\ref{fig:arch}, our pipeline starts from a cold-start model obtained via supervised fine-tuning on the GETok vocabulary. This process provides the policy \(\pi_{\theta}\) with a foundation for generating spatially grounded responses in GETok. The training then follows a two-stage procedure utilizing GRPO~\cite{shao2024deepseekmath}. The first stage focuses on generating grid tokens, rewarding spatial accuracy and structural validity, while the second stage introduces offset tokens in multi-turn dialogues, incentivizing precision improvement through iterative refinement. This self-correcting mechanism significantly enhances localization precision while ensuring conversational coherence.

We design specific reward functions for grid and offset tokens. Grid token rewards promote accurate regional grounding, while offset token rewards aim to adjust misaligned grid anchors through local shifts. This dual-phase optimization distinctly separates token placement from token movement, achieving geometry-aware self-correction.

\subsubsection{Reward for Grid Token Generation}
The key distinction from existing methods is that we train the model to generate a variable-size set of semantic-critical points to handle complex scenes, rather than reducing each mask to only one or two points. Specifically, the reward for grid-token generation is designed as follows:

\noindent \textbf{Format Reward.}
This reward encourages structured output with reasoning in \texttt{<think>} tags and spatial predictions in \texttt{<answer>} tags containing \texttt{<box>} and \texttt{<seg>} tokens.

\noindent \textbf{Non-repeat Reward.}
This reward penalizes sentence-level repetition when multiple identical sentences appear.

\noindent \textbf{Mask Reward.} 
This reward encourages higher IoU scores between the generated masks and ground-truth masks using a piecewise function. For automated quality assessment, we employ SAM~\cite{sam} and convert predicted boxes and points into spatial prompts to generate masks. 

\noindent\textbf{Box Reward.}
This reward encourages higher box IoU scores and smaller L1 corner distances between the predicted and ground-truth box corners. 

\noindent\textbf{Semantic-Critical Points Reward.}
This reward evaluates segmentation quality by combining the hit ratio (points inside ground-truth masks) and spatial distribution. We balance point density using an exponential saturation term $(1-e^{-m_p/5})$ to prevent sparse predictions and a linear penalty $(0.02m_p)$ to avoid excessive points. 

\subsubsection{Reward for Offset Token Refinement}
\label{sec:offset-reward}
In our experiments, poorly designed reward formulations often lead the model to predict no offsets at all. To avoid this collapse, we design a set of refinement rewards that explicitly encourage meaningful geometric updates.

\noindent\textbf{Format Reward.} 
This reward enforces a minimal schema on per-instance \texttt{<offset>} tokens containing \texttt{<box>} and \texttt{<seg>} serializations, as we find that the \texttt{<think>} preamble provides negligible benefit for offsets.

\noindent\textbf{Point Refinement Reward.}
This reward assesses point-level refinement with a ternary score $s_{k,p} \in \{-1,0,1\}$ per point: $-1$ for moves that leave the ground-truth mask, $+1$ for corrections that enter the mask, stay inside it, or perform a valid deletion, and $0$ otherwise. A deletion is counted as valid only when \del\ is predicted and no point in the $3\times3$ neighborhood of the original position lies inside the ground-truth mask. 

\noindent\textbf{Box Refinement Reward.} This reward measures the IoU gain between the initial and refined bounding boxes. For each instance, we assign a positive reward when the refined box increases the bounding box IoU over the initial prediction, and zero otherwise.

\noindent\textbf{Mask IoU Gain Reward.} This reward favors larger IoU improvements by measuring the relative gain, defined as the IoU improvement from the initial proposal to the refined result, normalized by the maximum possible improvement.

\section{Experiments}
\begin{table*}[t]
\centering
\setlength{\tabcolsep}{12pt} 
\renewcommand{\arraystretch}{0.95}
\caption{\textbf{Referring Expression Comprehension} (REC) results on the RefCOCO (+/g) datasets.} 
\label{tab:rec}
\resizebox{0.96\textwidth}{!}{ 
\small
\begin{tabular}{lccccccccccccc}
\toprule
\multirow{2}*{\textbf{Methods}} & & \multicolumn{3}{c}{\textbf{refCOCO}} && \multicolumn{3}{c}{\textbf{refCOCO+}} && \multicolumn{2}{c}{\textbf{refCOCOg}} && \multirow{2}*{\textbf{Avg.}}\\
\cmidrule(lr){3-5} \cmidrule(lr){7-9} \cmidrule(lr){11-12}
& & \textbf{Val.} & \textbf{Test-A} & \textbf{Test-B} && \textbf{Val.} & \textbf{Test-A} &\textbf{ Test-B} && \textbf{Val.} & \textbf{Test} && \\
\midrule
\midrule
\multicolumn{14}{c}{\textit{--Supervised Fine-Tuning Models \textbf{(Acc@0.5)}--}}\\
UNINEXT-L \cite{yan2023universal} && 91.4 & 93.7 & 88.9 && 83.1 & 87.9 & 76.2 && 86.9 & \textbf{87.5} && 87.0 \\
Shikra \cite{chen2023shikra} && 87.0 & 90.6 & 80.2 && 81.6 & 87.4 & 72.1 && 82.3 & 82.2 && 82.9\\
Ferret \cite{you2023ferret} && 87.5 & 91.4 & 82.5 && 80.8 & 87.4 & 73.1 && 83.9 & 84.8 && 83.9 \\
Groma \cite{ma2024groma} && 89.5 & 92.1 & 86.3 && 83.9 & 88.9 & 78.1 && {86.4} & {87.0} && 86.5 \\ 
ClawMachineX \cite{ma2024clawmachine} && 89.7 & 92.5 & 86.9 && 84.4 & 88.9 & 78.0 && {86.7} & {87.1} && 86.8 \\ 
Qwen2.5-VL-7B~\cite{bai2025qwen2} && 90.0 &92.5 & 85.4&& 84.2 & 89.1 & 76.9 && {87.2} & 87.2 && 86.6 \\ 
\rowcolor{violet!6}{GETok-SFT-grid} && {90.4} &\textbf{93.8} & {86.9} &&{86.3} &{90.8} & {79.4} && {87.1} & {87.5} && {87.8} \\ 
\rowcolor{violet!6}{GETok-SFT} && \textbf{90.6} &{93.7} & \textbf{87.2} &&\textbf{86.7} &\textbf{90.9} & \textbf{79.9} && \best{88.5} & \textbf{88.4} && \textbf{88.2} \\ 
\hdashline[2pt/2pt]
\multicolumn{14}{c}{\textit{--Supervised Fine-Tuning Models \textbf{(Acc@0.8)}--}}\\
Qwen2.5-VL-7B~\cite{bai2025qwen2} && 74.0  &78.6 &67.2 &&68.8  &75.4  &59.4  && {71.1} & 72.3 && 70.9\\
\rowcolor{violet!6}{GETok-SFT-grid} &&{74.1}  &{78.8} & {68.5} &&{69.4} &{75.9} &{61.8}  &&{72.4}  &{73.5}  &&{71.8}  \\ 
\rowcolor{violet!6}{GETok-SFT} &&\textbf{74.8}  &\textbf{79.6} & \textbf{69.8} &&\textbf{70.3} &\textbf{77.7} &\textbf{62.9}  &&\textbf{72.8}  &\textbf{74.1}  &&\textbf{72.8}  \\ 
\midrule
\multicolumn{14}{c}{\textit{—— Reinforcement Learning Models \textbf{(Acc@0.5)} ——}}\\
VisionReasoner$^{\dag}$ \cite{liu2025visionreasoner} && 89.6 &91.1 & -&& 85.4 & 89.0 & - && 88.2 & {89.0} && 88.7 \\ 
\rowcolor{violet!6}{GETok-R1-grid} && {90.2} &{92.9} & - &&{86.7} &{89.9} & {-} && {89.2} & {88.7} && {89.6} \\ 
\rowcolor{violet!6}{GETok-R1} && \textbf{90.9} &\textbf{93.6} & - &&\textbf{87.1} &\textbf{90.8} & \textbf{-} && \textbf{89.9} & \textbf{89.2} && \textbf{90.3} \\ 
\hdashline[2pt/2pt]
\multicolumn{14}{c}{\textit{—— Reinforcement Learning Models \textbf{(Acc@0.8)} ——}}\\
VisionReasoner$^{\dag}$ \cite{liu2025visionreasoner} &&74.2  & 78.4& -&&68.9 &75.3  & - &&72.1  & 73.2 &&73.7 \\ 
\rowcolor{violet!6}{GETok-R1-grid} && {74.8} &{78.9} & -&&{70.1} &{76.6} & {-} && {73.6} & {74.9} && {74.8} \\ 
\rowcolor{violet!6}{GETok-R1} && \textbf{75.6} &\textbf{80.1} & -&&\textbf{71.6} &\textbf{78.2} & {-} && \textbf{74.8} & \textbf{76.1} && \textbf{76.1} \\ 
\hline
\end{tabular}}
\end{table*}

\subsection{Experimental Setup}
\label{sec:setting}
\noindent\textbf{Training Details.} We use Qwen2.5-VL-7B~\cite{bai2025qwen2}, a powerful open-source VLM, as the base model for GETok.
For GETok-SFT, we use the \texttt{ms\_swift} framework~\cite{msswift}  with LoRA~\cite{hu2021lora} (rank=64), a batch size of 16, and a learning rate of $1\times10^{-6}$, training on publicly available corpora spanning image-level reasoning, referring grounding, and segmentation. 
For GETok-RL, we employ the GRPO algorithm~\cite{shao2024deepseekmath} via the \texttt{easy-r1} framework~\cite{zheng2025easyr1}, initializing from a cold-start model trained on referring segmentation data and open-source multimodal instruction data (e.g., LLaVA-CoT-100k~\cite{xu2025llavacot}). GRPO training in stage 1 uses a 9K dataset composed of LISA++~\cite{yang2023lisa++} and referring segmentation samples~\cite{yu2016refcoco, mao2016refcocog}, with a batch size of 16 (8 samples per step), learning rate of $1\times10^{-6}$, and weight decay of 0.01. Refinement training in stage 2 is limited to 200 steps to prevent overfitting, given the concise nature of offset tokens. 
All experiments are conducted on 8×80GB GPUs using the DeepSpeed engine~\cite{rasley2020deepspeed}, with a grid size of 32 and an offset size of 64. 

\noindent\textbf{Benchmark Settings.}
GETok addresses a broad spectrum of visual referring tasks. 
We conduct quantitative evaluations on eight settings:
(i) Referring Expression Comprehension (REC),
(ii) Referring Expression Segmentation (RES),
(iii) Reasoning Segmentation,
(iv) Referring Captioning,
(v) Generalized Referring Expression Segmentation (gRES),
(vi) Lane Polyline Detection,
and (vii) Object Pointing.
We also build (viii) a driving case study that mixes polylines (lanes), polygons (drivable areas), and boxes (dynamic objects), demonstrating unified supervision in complex scenes.
For GETok-SFT, we perform extensive validation across all eight settings (i)–(viii), establishing strong and consistent SFT baselines under a shared training setting and decoding budget. 
For GETok-RL, we focus on (i)–(iii), which reflect mainstream benchmarks for R1-style referring models. Due to space limitations, complete results and ablation studies are provided in the supplementary material.

\begin{figure*}[t]
    \centering
    \includegraphics[width=\linewidth]{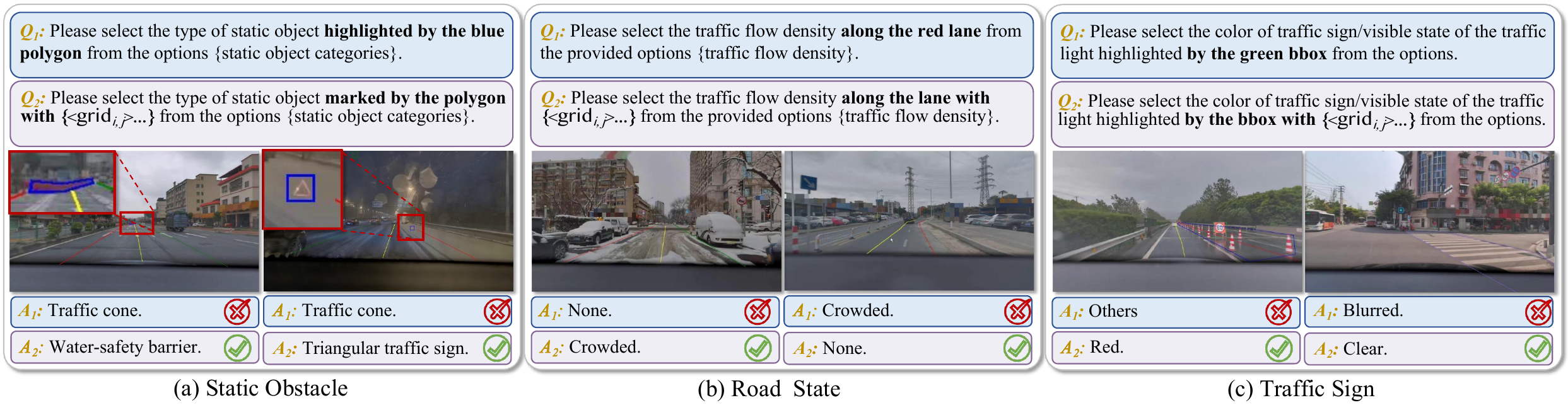}
    \caption{
    \textbf{Qualitative results of applying the proposed grid tokens to driving scene.} 
    Challenging examples from three referring categories demonstrate that the proposed GETok offers superior region-referencing ability compared to conventional visual referring prompts.
    }
    \label{fig:driving}
    \vspace{-5pt}
\end{figure*}
\subsection{Overall Performance}
\label{sec:main_result}
\noindent \textbf{Referring Expression Segmentation.} 
As shown in Tab.~\ref{tab:res}, GETok-SFT demonstrates competitive performance compared to specialized methods while maintaining architectural simplicity. When trained with our reinforcement learning framework, GETok-RL achieves state-of-the-art performance, fully realizing the potential of our token design with a significant gain of +4.5\% over supervised fine-tuning. This highlights the substantial capability of our regularized 2D token representation in RL paradigms, where the structured action space facilitates stable policy optimization and efficient exploration.

The offset mechanism proves essential in both training paradigms, providing consistent gains of +1.0\% in SFT and +1.5\% in RL over grid-only configurations. This improvement is particularly critical for mask generation tasks, where even minor localization errors can be amplified during the decoding process, highlighting the importance of precise spatial refinement.

Fig.~\ref{fig:visual1}(a) shows that using off-the-shelf SAM allows us to retain SAM’s generalization capability, resulting in high-quality masks with fine-grained edge details. We note that this can sometimes lead to discrepancies when compared to lower-quality ground truth annotations. Figs. \ref{fig:visual1}(b) and (d) demonstrate the adaptability of our refinement mechanism, which applies small corrections to accurate proposals (b) and larger corrections to less precise ones (d). Fig.~\ref{fig:visual1}(c) specifically showcases the effectiveness of our \textit{propose-and-refine} approach for small targets, where precise localization is particularly challenging.

\vspace{1mm}
\noindent \textbf{Referring Expression Comprehension.} 
As indicated in Tab. \ref{tab:rec}, GETok-SFT demonstrates solid performance under the conventional accuracy metric (Acc@0.5), with a gain of +1.6\% over the Qwen2.5-VL-7B baseline. 
To better evaluate localization accuracy, we report results using the stricter Acc@0.8 metric. 
Under this stricter evaluation, the combination of grid and offset tokens shows significant improvements in spatial reasoning. The visualizations reveal particularly pronounced gains for small objects under both the SFT and RL settings.
Unlike the ReasonSeg dataset~\cite{lisa}, which comprises complex reasoning chains, RefCOCO expressions are relatively straightforward, leaving less room for RL-based refinement. This contrast highlights that our GETok-RL achieves the greatest advantages when tackling complex reasoning tasks that benefit from iterative refinement and chain-of-thought reasoning.

\begin{table}[t]
    \centering
    \caption{{Performance comparison of different grid resolutions on REC (Acc@0.8) and RES (gIoU).} }
    \setlength{\tabcolsep}{14pt} 
    \renewcommand{\arraystretch}{1}
    \resizebox{0.9\linewidth}{!}{
    \begin{tabular}{cccc}
        \toprule[1pt]
        \textbf{Grid Size} & \textbf{REC} & \textbf{RES} & \textbf{Avg. Token Len. per Mask} \\
        \midrule
        \midrule
        $16 \times 16$ & 68.9 & 66.2  & 5.2\\
        $64 \times 64$ & 71.2 & 67.1 & 14.6\\
        \midrule
        \rowcolor{violet!6}$32 \times 32$ & 70.9 & 67.2 & 8.7\\
        \rowcolor{violet!6}w/ offset & \textbf{72.6} & \textbf{68.2} & 9.2\\

        \bottomrule[1pt]
    \end{tabular}
    }
    \label{tab:gridsize}
    \vspace{-5pt}
\end{table}
\subsection{Grid Resolution}
The grid size $n$ is a crucial parameter for GETok, governing the trade-off between spatial precision and vocabulary expansion. As shown in Tab.~\ref{tab:gridsize}, we identify two key observations:
First, the $32 \times 32$ configuration achieves comparable performance to $64 \times 64$ while maintaining significantly lower token length and vocabulary overhead.
Second, offset tokens demonstrate remarkable efficiency, outperforming the costly doubling of grid resolution with only 10 additional tokens. This minimal expansion delivers superior performance compared to the $64 \times 64$ configuration.

\subsection{Real-World Driving Case Study}
\label{sec:maindriving}
We further evaluate grid tokens using a proprietary driving dataset that features diverse urban scenarios, annotated in three ways: lanes (polylines), static obstacles (bounding boxes), and traffic signs (key points). More details can be found in the supplementary materials.
For general scene understanding, GETok consistently outperforms traditional visual prompts across all tasks, achieving significant improvements in challenging scenarios: a +12.24\% increase in traffic sign color recognition and a +7.95\% increase in static obstacle classification, as shown in Tab.~\ref{tab:driving}.
%
Fig.~\ref{fig:driving} illustrates the success of GETok in complex driving scenarios, demonstrating its ability to handle diverse reference types through a unified representation without requiring architectural modifications. 
Additionally, we report lane detection results for GETok, highlighting its particular strength in handling curved lanes. For lane detection, GETok transforms continuous coordinate regression into discrete point selection, resulting in a +3\% increase in precision, a +18\% increase in recall, and a +10\% increase in F1-score compared to coordinate-based methods, as shown in Tab.~\ref{tab:lane}.
\begin{table}[t]
\centering
\renewcommand{\arraystretch}{1}
\caption{Performance comparison of using GETok with the visual referring prompt in the driving scene.}
\label{tab:driving}
\setlength{\tabcolsep}{13pt} 
\resizebox{0.9\linewidth}{!}{
\begin{tabular}{llc>{\columncolor{violet!6}}c}
\toprule
\textbf{Category}&\textbf{Task} & \textbf{Baseline} & \textbf{GETok}  \\
\midrule
\midrule
\multirow{2}{*}{\textbf{Static obstacle}}& Classification & 81.69 & \textbf{89.64} \\
 & Visible State & 90.60 & \textbf{93.49} \\
\midrule
\multirow{3}{*}{\textbf{Road}}& Blockage Status & 86.07 & \textbf{87.25} \\
 & Surface Condition & 95.46 & \textbf{95.68}\\
 & Traffic Density & 84.31 & \textbf{86.39}\\
\midrule
\multirow{2}{*}{\textbf{Traffic Sign}}& Color & 71.43 & \textbf{83.67} \\
 & Visible State & 63.27 & \textbf{67.35}\\
\bottomrule
\end{tabular}
}
\vspace{2pt}
\end{table}

\begin{table}[t]
\centering
\setlength{\tabcolsep}{18pt} 
\renewcommand{\arraystretch}{1}
\caption{Comparative results for lane polyline detection.}
\label{tab:lane}
\resizebox{0.9\linewidth}{!}{
\begin{tabular}{cccc}
\toprule
\multirow{2}*{\textbf{Methods}}&\multicolumn{3}{c}{\textbf{Lane Polyline}}\\
\cmidrule{2-4}
 & \textbf{Precision} & \textbf{Recall} & \textbf{F1 Score} \\
\midrule
\midrule
Coords-based & 0.49 & 0.47 & 0.48\\
\rowcolor{violet!6}GETok & \textbf{0.52} & \textbf{0.65} & \textbf{0.58} \\
\bottomrule
\end{tabular}
}
\vspace{-5pt}
\end{table}
\section{Conclusion}
We presented GETok, a novel spatial representation that addresses the fundamental challenge of 2D spatial reasoning in MLLMs. By introducing learnable grid and offset tokens, GETok provides a unified framework for precise spatial localization while maintaining architectural simplicity.
The offset mechanism yields the emergent benefit of progressive localization refinement, enabling iterative self-correction.
Extensive experiments across diverse referring tasks show that GETok is effective under both supervised fine-tuning and reinforcement learning, while retaining a simple and unified autoregressive formulation.

\newpage
\paragraph{Acknowledgements.} This work was supported in part by NSFC (62322113, 62376156), Shanghai Municipal Science and Technology Major Project (2025SHZDZX025G15, 2021SHZDZX0102), and the Fundamental Research Funds for the Central Universities.

{
    \small
    \bibliographystyle{ieeenat_fullname}
    \bibliography{main}
}

\appendix
\renewcommand{\thesection}{\Alph{section}}
\renewcommand{\thetable}{\Alph{table}}
\renewcommand{\thefigure}{\Alph{figure}}

\clearpage
\setcounter{page}{1}
\maketitlesupplementary

\noindent We provide supplementary material for further study and analysis related to the main paper, arranged as follows:
\begin{itemize}
\item Additional experimental results extending the main findings (Sec.~\ref{app:training results})
\item Real-world driving dataset curation (Sec.~\ref{app:driving})
\item Additional implementation details, including training setup, offset-aware dataset construction, and reward design (Sec.~\ref{app:implementation_details})
\item Additional qualitative results and visual analysis (Sec.~\ref{app:visualization_results})
\end{itemize}

\section{Additional Experimental Results}
\label{app:training results}

\subsection{More Benchmarks}

\vspace{5pt}
\noindent\textbf{Referring Captioning} evaluates region understanding given referring inputs (e.g., bounding boxes, masks).
We evaluate region-based caption generation on refCOCOg~\cite{mao2016refcocog} and Visual Genome~\cite{VG}. As shown in Tab.~\ref{tab:regioncap}, GETok achieves competitive or superior performance relative to models that rely on specialized region feature extractors (\checkmark), highlighting the effectiveness of GETok for region-aware comprehension. 
GETok is particularly effective in scenarios with overlapping objects, where traditional bounding boxes often fail to precisely capture targeted regions. 

\begin{table}[htbp]
\centering
\caption{\textbf{Region-Level Captioning} results on the refCOCOg and visual genome datasets.}
\label{tab:regioncap}
\setlength{\tabcolsep}{5pt} 
\renewcommand{\arraystretch}{1}
\resizebox{0.9\linewidth}{!}{
\begin{tabular}{lccccc}
\toprule
\multirow{2}{*}{\textbf{Methods}} & \multirow{2}{*}{\makecell{\textbf{Region Feat.}\\\textbf{Extractor}}} & \multicolumn{2}{c}{\textbf{refCOCOg}} & \multicolumn{2}{c}{\textbf{Visual Genome}} \\ 
\cmidrule(lr){3-4} \cmidrule(lr){5-6} 
& & METEOR & CIDEr & METEOR & CIDEr \\ 
\midrule
\midrule
GRIT \cite{wu2024grit} & \ding{52} & 15.2 & 71.6 & 17.1 & 142.0 \\
SLR \cite{SLR} & \ding{52} & 15.9 & 66.2 & - & - \\
GPT4RoI \cite{zhang2023gpt4roi} & \ding{52} & - & - & 17.4 & 145.2 \\
GLaMM \cite{glamm} & \ding{52} & 16.2 & 106.0 & 19.7 & \textbf{180.5} \\
Groma \cite{ma2024groma} & \ding{52} & 16.8 & 107.3 & 19.0 & 158.4 \\
Kosmos-2 \cite{peng2023kosmos} & \ding{56} & 14.1 & 62.3 & - & - \\
Shikra-7B \cite{chen2023shikra} & \ding{56} & 15.2 & 72.7 & - & - \\
\rowcolor{violet!6}\textbf{GETok-SFT} & \ding{56} & \textbf{16.9} & \textbf{110.5} & \textbf{19.0} & 165.9 \\
\bottomrule
\end{tabular}
}
\end{table}

\vspace{5pt}
\noindent\textbf{Generalized RES} validates multi-instance grounding through grid token sequences, demonstrating simultaneous referencing capability for multiple objects within a single spatial representation.
GETok naturally supports multi-instance expressions. 
We evaluate GETok on the gRefCOCO dataset~\cite{liu2023gres} for multi-instance segmentation. As shown in Tab.~\ref{tab:gres}, GETok achieves competitive performance relative to specialized methods while maintaining architectural simplicity.

\begin{table}[htbp]
\centering
\caption{\textbf{Generalized Referring Expression Segmentation} results (cIoU) on the RefCOCO (+/g) datasets.}
\label{tab:gres}
\setlength{\tabcolsep}{6pt} 
\renewcommand{\arraystretch}{1}
\resizebox{0.9\linewidth}{!}{
\begin{tabular}{lcccccccc}
\toprule
\multirow{2}*{\textbf{Methods}}&  \multirow{2}*{\makecell{\textbf{Training }\\\textbf{M-Dec.}}} & \multirow{2}*{\textbf{Validation}} && \multirow{2}*{\textbf{Test-A}} && \multirow{2}*{\textbf{Test-B}} && \multirow{2}*{\textbf{Average}}\\
&&&&&&&&\\
\midrule
\midrule
LAVT \cite{yang2022lavt} &\ding{52} & 58.4  && 65.9  && 55.8  && 60.0\\
ReLA \cite{liu2023gres}&\ding{52} &  63.6  && 70.0 && 61.0 && 64.9 \\
LISA \cite{lisa}&\ding{52} & 63.5 && 68.2 && 61.8 && 64.5\\
GSVA \cite{gsva}&\ding{52} & 68.0  && 71.8  && 63.8 && 67.9 \\
\rowcolor{violet!6}\textbf{GETok-SFT}&\ding{56}& {66.9}  && {72.3}  && {64.1}  && {67.8}  \\
\rowcolor{violet!6}\textbf{GETok-RL}&\ding{56}& \textbf{67.4}  && \textbf{74.1}  && \textbf{65.6}  && \textbf{69.0}  \\
\bottomrule
\end{tabular}
}
\end{table}

\vspace{7pt}
\noindent \textbf{Object Pointing} evaluates precise point-level localization. Instead of restricting the target to a single predefined point, GETok supports flexible point annotations by marking representative object positions, making the representation more adaptable to diverse object types and scene layouts.
As shown in Tab.~\ref{tab:pointing}, GETok achieves competitive performance compared to methods trained with substantially more data. The advantage is particularly pronounced in dense object scenarios, where grid tokens reduce coordinate representation from multiple sequential tokens (e.g., \texttt{['(', '124', ',', '143', ')']}) to \emph{a single} spatial token (e.g., \texttt{<grid$_{12,14}$>}), eliminating the formatting errors that accumulate with longer text-based coordinate sequences.

\begin{table}[htbp]
\setlength{\tabcolsep}{8pt} 
\caption{\textbf{Object Pointing} results on HumanRef and RefCOCOg datasets.}
    \label{tab:pointing}
    \centering
    \resizebox{0.9\linewidth}{!}{
       \begin{tabular}{cccc}
\toprule
\textbf{Methods} & \textbf{HumanRef}      & \textbf{refCOCOg val}  & \textbf{refCOCOg test} \\ 
\midrule
\midrule
OVIS2.5-9B~\cite{lu2024ovis}              & 62.3          & \textbf{85.0} & \textbf{{84.5}} \\
Molmo-7B-D~\cite{deitke2025molmo}               & 70.0          & 83.7          & 83.6          \\
Qwen2.5-VL-7B~\cite{bai2025qwen2}            & 65.1          & 78.9          & 79.4          \\
\rowcolor{violet!6}\textbf{GETok-SFT}  & \textbf{70.7} & {{84.1}} & {82.9} \\ 
\bottomrule
\end{tabular}
    }
\end{table}

\begin{figure*}[htbp]
    \centering
    \includegraphics[width=0.96\linewidth]{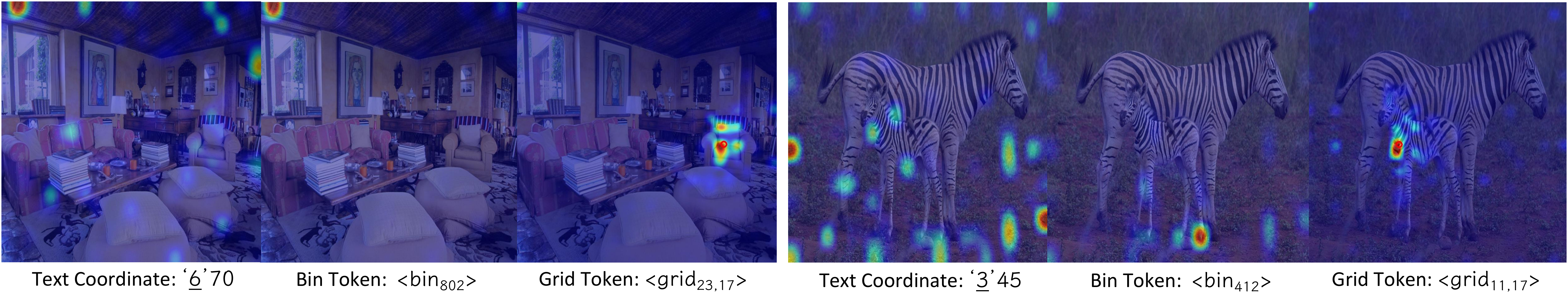}
    \caption{\textbf{Visualization of spatial responses for different localization vocabularies.} We aggregate attention maps between location tokens and image patches to obtain heatmaps for text coordinates, 1D bin tokens, and grid tokens. Grid tokens produce smooth, topology-aware activations that align with object extents.}
    \label{fig:heatmap}
\end{figure*}

\subsection{More Discussions}

\noindent\textbf{How Should Points be Represented?} 
We analyze three representation formats that operate purely through \emph{vocabulary-level modifications}: text coordinates, bin tokens, and grid tokens, all of which require no architectural changes. Among them, bin tokens and text coordinates share the same 1D numerical nature, with bin tokens merely quantizing coordinates into discrete indices, and empirical evidence shows that bin-based methods can even underperform text coordinates~\cite{chen2023shikra}. The key difference, therefore, lies between these 1D schemes and the 2D spatial encoding of grid tokens, which addresses three fundamental limitations:

\vspace{2pt}
\noindent\emph{1) 1D-2D Representation Gap:}
A single 1D token cannot directly represent a 2D location; instead, multiple tokens must be combined to denote a coordinate. This composition hinders the implicit semantic features of the 2D space from being effectively mapped into the token embeddings.

\noindent \emph{2) Format Brittleness:} Syntactic elements introduce exponential failure rates that are particularly problematic in multi-object scenarios. For example, with 98\% per-token accuracy, a 12-token box sequence has a 78\% validity probability, dropping to 48\% for three boxes (36 tokens).

\noindent \emph{3) Metric–Objective Mismatch:} Token-level cross-entropy on digit sequences correlates poorly with geometric error. Small changes in token indices can correspond to large jumps in image space.

Using Qwen2.5-VL-7B with identical RefCOCO/+/g instruction-tuning data, we compare text, bin, and grid formats in Tab.~\ref{tab:format}, and observe a clear advantage for grid tokens. Furthermore, as shown in Fig.~\ref{fig:heatmap}, grid tokens produce smooth, locally coherent activations that closely follow object extents because each token is tied to a fixed 2D region in the image plane. In contrast, text and bin tokens yield fragmented, geometry-agnostic responses without a stable 2D correspondence. 

\begin{table}[htbp]
\centering
\caption{Ablation on \textbf{point representation formats} for REC on the RefCOCO/+/g datasets.}
\label{tab:format}
\setlength{\tabcolsep}{4pt} 
\resizebox{0.9\columnwidth}{!}{%
\begin{tabular}{lccc}
\toprule
\textbf{Methods} & \textbf{refCOCO Test-A} & \textbf{refCOCO+ Test-A} & \textbf{refCOCOg Test} \\
\midrule
\midrule
Text Coordinates &92.9 &89.9 &87.4\\
Bin tokens &92.3 &89.9 &87.1\\
\rowcolor{violet!6}Grid tokens & \textbf{93.0} &\textbf{90.6} &\textbf{87.6}\\
\bottomrule
\end{tabular}%
}
\end{table}

\begin{figure}[htbp]
    \centering
    \includegraphics[width=0.96\linewidth]{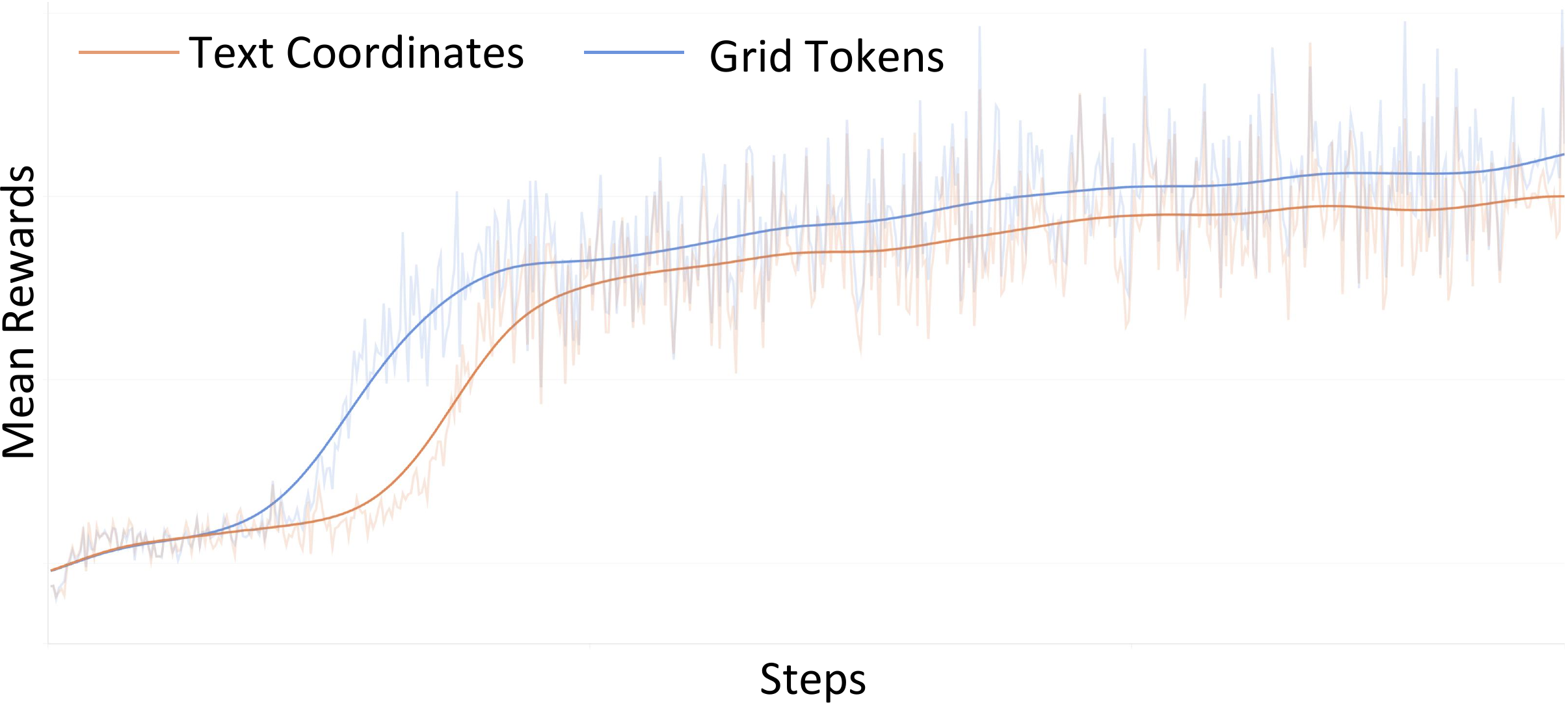}
    \caption{\textbf{Reward curve comparison between grid tokens and text coordinates.} GETok achieves faster convergence and higher rewards than text coordinates.}
    \label{fig:reward}
\end{figure}

\noindent\textbf{Why GRPO Works with GETok?}
GETok's structured representation creates an ideal action space for GRPO optimization. 
As shown in Fig.~\ref{fig:reward}, GETok achieves accelerated convergence at equivalent training steps compared to text coordinates, validating its structured action space advantage for GRPO optimization. We attribute this advantage to two key factors: 
(1) The 2D grid structure provides a stable foundation for policy learning, unlike text coordinates, where minor token changes yield discontinuous spatial shifts.
(2) The finite $n \times n$ token format is easier to learn than text coordinates. This compact set allows the model to focus on spatial layout rather than complex text patterns, leading to faster convergence.

\begin{figure*}[t]
    \centering
    \includegraphics[width=0.96\linewidth]{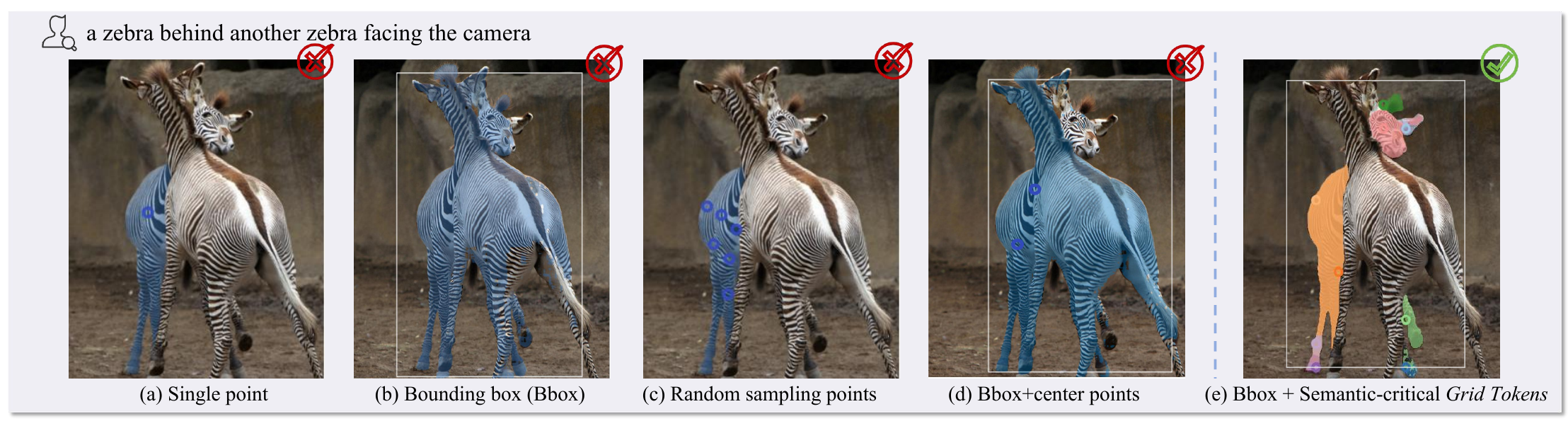}
    \caption{\textbf{Comparison of mask representation strategies.} We convert continuous masks into discrete, segment-critical grid tokens to achieve precise region referencing.}
    \label{fig:pointtoken}
\end{figure*}

\vspace{5pt}
\noindent\textbf{How to Represent Masks with Sparse Geometry?} 
\label{sup:mask}
We analyze existing sparse geometric representations, such as single points, bounding boxes, fixed sets of one or two points, or randomly sampled points, all of which suffer from redundancy and an inability to unambiguously capture complex mask semantics as shown in Fig.~\ref{fig:pointtoken}.
We introduce a novel greedy algorithm that automatically extracts an appropriate set of such tokens from a target mask. 
Compared to methods that require training a dedicated mask decoder~\cite{lisa,gsva,glamm}, this design offers several advantages:

\noindent \emph{1) At training time}, our method avoids any mask-specific loss, decoder, or supervision, offering a simpler alternative compared to methods that rely on task-specific decoders. 

\noindent \emph{2) At inference time}, our method offers strong flexibility as our decoder is purely plug-and-play and can be seamlessly updated without retraining the referring VLM. 
For example, replacing SAM~\cite{sam} with advanced SAM2~\cite{ravi2024sam2}, our method achieves a performance gain of 0.8\% cIoU on refCOCO val at no cost. In contrast, LISA has to retrain the full model for this replacement, which is particularly costly.

\subsection{Ablation Studies}
\vspace{5pt}
\noindent\textbf{Image Preprocessing.}
We investigate the impact of different image preprocessing strategies on localization performance as shown in Tab.~\ref{tab:preprocess}. Padding gives the worst results, because the added gray borders effectively downscale the informative region and distract the model from relevant content. Center cropping risks semantic distortion by removing peripheral image areas. For example, in a referring expression such as ``the person on the far left," cropping may exclude the target entirely, leading to ground-truth mismatch. 
In contrast, resizing and dynamic resolution achieve comparable performance in our experiments. We therefore adopt simple resizing as our default preprocessing strategy.
\begin{table}[h]
\centering
\caption{Ablation on \textbf{image preprocessing} strategies for REC on RefCOCOg.}
\label{tab:preprocess}
\setlength{\tabcolsep}{22pt} 
\resizebox{0.6\columnwidth}{!}{%
\begin{tabular}{lc}
\toprule
\textbf{Methods} & \textbf{RefCOCOg} \\
\midrule
\midrule
Padding & 85.9 \\
Center Crop & 86.2\\
Dynamic & {87.1} \\
\rowcolor{violet!6}Resize & \textbf{87.4} \\
\bottomrule
\end{tabular}%
}
\end{table}

\vspace{5pt}
\noindent\textbf{Reward Function.}
For grid token generation, removing the semantic-critical points reward causes the model either to collapse to one or two high-confidence points or to overpopulate a small region with redundant points, as shown in Tab.~\ref{tab:abl_reward_all}. Removing the box reward yields the largest drop, and visual inspection shows that points become scattered in the absence of a stable coarse prior. By contrast, the mask reward mainly provides fine-grained geometric supervision, especially for thin structures and concave regions that are not well constrained by box and point-level signals alone.

For offset token refinement, we focus on whether offsets perform genuine geometric corrections. The mask IoU gain and box refinement rewards provide instance-level guidance that promotes updates with improved mask and box IoU. The point refinement reward further stabilizes behavior by reducing large mask changes caused by a few erroneous point adjustments.

\begin{table}[htbp]
\centering
\setlength{\tabcolsep}{4pt}
\caption{Ablation on \textbf{reward design} for grid-token generation and offset-token refinement.}
\label{tab:abl_reward_all}
\resizebox{0.9\linewidth}{!}{
\begin{tabular}{lcccc}
\toprule
\multicolumn{5}{c}{\textbf{Reward for Grid Token Generation}} \\
Variant 
& Mask & Box & Sem. points & ReasonSeg \\
\midrule
w/o Sem. points           & \ding{52} & \ding{52} &           & 58.6 \\
w/o Mask reward           &           & \ding{52} & \ding{52} & 59.1 \\
w/o Box reward            & \ding{52} &           & \ding{52} & 57.2 \\
\rowcolor{violet!6}Full (ours)               & \ding{52} & \ding{52} & \ding{52} & \textbf{60.1} \\
\midrule
\multicolumn{5}{c}{\textbf{Reward for Offset Token Refinement}} \\
Variant 
& Point gain & Box gain & Mask IoU gain & ReasonSeg \\
\midrule
w/o Mask IoU gain    & \ding{52} & \ding{52} &           & 61.8 \\
w/o Box ref.         & \ding{52} &           & \ding{52} & 61.2 \\
w/o Point ref.       &           & \ding{52} & \ding{52} & 60.5 \\
\rowcolor{violet!6}Full (ours)               & \ding{52} & \ding{52} & \ding{52} & \textbf{62.8} \\
\bottomrule
\end{tabular}
}
\end{table}

\vspace{5pt}
\noindent\textbf{Reasoning vs.\ No Reasoning for Offset Refinement.}
The \texttt{<think>} process has been shown to be beneficial for multimodal understanding, especially in cases that require complex semantic reasoning~\cite{liu2025segzero,liu2025visual,shen2025vlmr1}. We further examine its role in the refine stage. Empirically, the performance gap between using and omitting \texttt{<think>} during refinement is negligible (0.1\% gIoU), suggesting that offset refinement does not substantially benefit from additional verbal reasoning. We observe that the model rarely produces meaningful explanations for point-level updates and instead repeats almost the same \texttt{<think>} content as in the propose step, so we do not enforce \texttt{<think>} generation in this stage.

\begin{figure*}[t]
    \centering
    \includegraphics[width=0.96\linewidth]{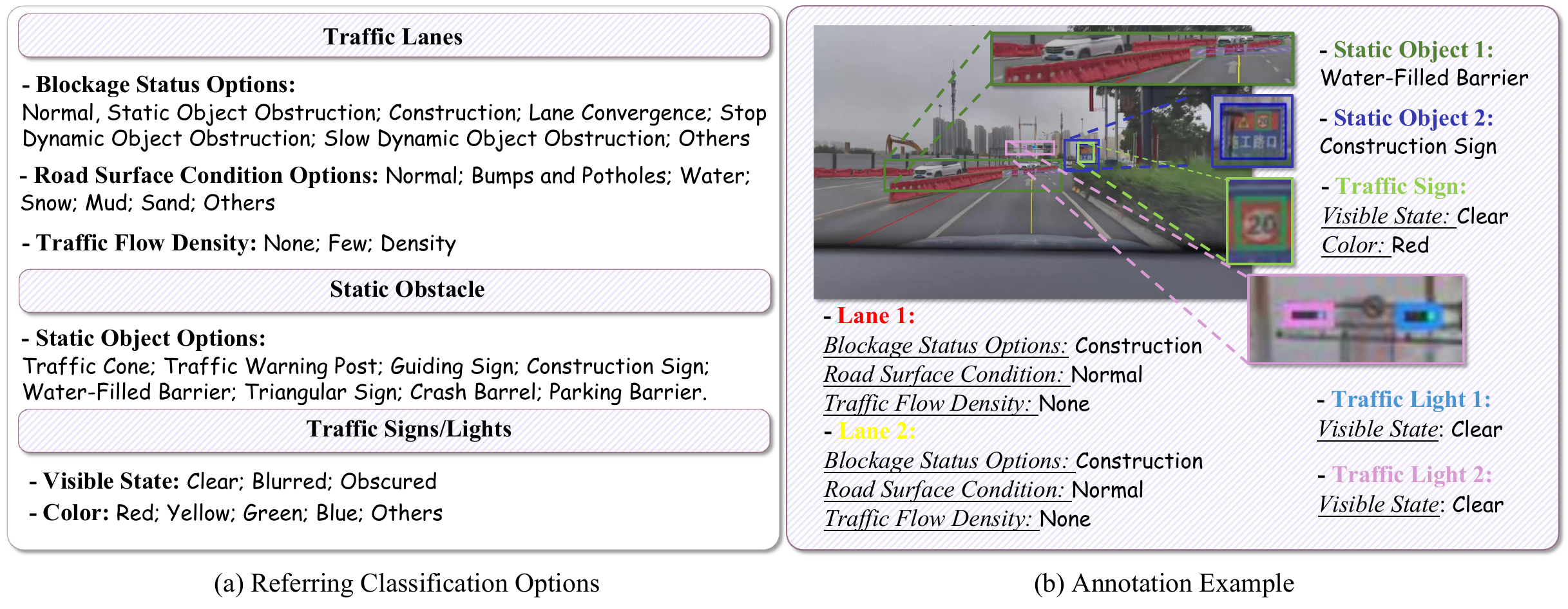}
    \caption{
    \textbf{Overview of the driving dataset annotations.} (a) Summarizes the taxonomy of annotated driving targets (lanes, static obstacles, and traffic signs/lights) with hierarchical labels. (b) Illustrates an example scene annotated with points, polygons, lane polylines, bounding boxes, and masks for referring and safety-related queries.
    }
    \label{fig:driving_example}
\end{figure*}

\section{Real-World Driving Dataset}
\label{app:driving}

We construct a proprietary autonomous driving dataset to evaluate GETok in complex real-world scenarios and to support comparison with strong baselines. The dataset contains 1,988 training samples with 29,825 annotations and 980 test samples with 14,524 annotations, covering diverse urban scenes including intersections, highways, and pedestrian zones.

As illustrated in Fig.~\ref{fig:driving_example}(a), the dataset categorizes driving targets into three classes: Traffic Lanes, Static Obstacles, and Traffic Signs with hierarchical annotations for multi-granular reasoning. Based on these annotations, we design a series of classification tasks to evaluate the model’s ability to understand and refer to specific regions in driving scenes.

Fig.~\ref{fig:driving_example}(b) shows an example from the dataset. Each sample is annotated using category labels selected from the taxonomy summarized in Fig.~\ref{fig:driving_example}(a). Overall, driving scenes provide a realistic setting that demands understanding and referring to regions in multiple formats, including points, polygons, polylines, bounding boxes, and masks, highlighting the application potential of a unified and robust localization framework.

\section{More Implementation Details}
\label{app:implementation_details}

\subsection{Training Setup}

\noindent\textbf{Supervised Fine-Tuning.} The model is fine-tuned on the mixed-task corpus summarized in Tab.~\ref{tab:training-data}. 
All location-related annotations (points, boxes, masks) are converted into GETok's grid tokens. 
The offset-aware dataset is constructed on top of RefCOCO/+/g, and a more systematic description of the construction pipeline is provided in Sec.~\ref{app:offset_dataset}.
We use a per-device batch size of 2 with 8 gradient accumulation steps, yielding an effective batch size of 16 per device.
All input images are resized to $840\times840$, and training is conducted with bfloat16 mixed precision.

\vspace{1mm}
\noindent\textbf{Reinforcement Learning.} We first perform a cold-start stage to adapt the model to the newly introduced tokens while mixing in CoT-style instruction data, thereby preserving its original multimodal capabilities.
Building on this checkpoint, we further optimize the policy with GRPO on both grid-token placement and offset-token refinement.
Each update is regularized by a KL-divergence penalty to the SFT policy with coefficient $1\times10^{-2}$.
For each prompt, we sample 8 candidate responses to estimate group-wise advantages.
For offset tokens, we empirically find that about 200 steps are sufficient to obtain satisfactory refinement performance.
\begin{table}[t]
\centering
\small
\caption{Summary of training data composition.}
\label{tab:training-data}
\setlength{\tabcolsep}{3pt}
\resizebox{\linewidth}{!}{%
\begin{tabular}{lll}
\toprule
\textbf{Stage} & \textbf{Datasets} & \textbf{Task} \\
\midrule
\multirow{8}{*}{\textbf{SFT}}
  & LLaVA-665K~\cite{liu2024llava} 
  & Image reasoning \\
  & \cellcolor{gray!15} RefCOCO/+/g~\cite{yu2016refcoco,mao2016refcocog} 
  & \cellcolor{gray!15} Referring grounding \\
  & COCO-Stuff~\cite{caesar2018coco}; ADE20K~\cite{zhou2017ade20k} 
  & Segmentation (seg.) \\
  & \cellcolor{gray!15} Visual Genome~\cite{VG} 
  & \cellcolor{gray!15} Image captioning \\
  & PACO-LVIS~\cite{ramanathan2023paco}; PASCAL-Part~\cite{chen2014detect} 
  & Part-level seg. \\
  & \cellcolor{gray!15} gRefCOCO~\cite{liu2023gres} 
  & \cellcolor{gray!15} Multi-instance seg. \\
  & Pixmo-point~\cite{deitke2025molmo} 
  & Object pointing \\
  & \cellcolor{gray!15} GETok-Offset 
  & \cellcolor{gray!15} Referring refinement \\
\midrule
\multirow{3}{*}{\textbf{\shortstack[c]{Cold\\Start}}}
  & RefCOCO/+/g~\cite{yu2016refcoco,mao2016refcocog} 
  & Referring seg. \\
  & \cellcolor{gray!15} LLaVA-CoT-100K~\cite{xu2025llavacot} 
  & \cellcolor{gray!15} Instruction tuning \\
  & GETok-Offset 
  & Offset training \\
\midrule
\multirow{2}{*}{\textbf{GRPO}}
  & RefCOCOg~\cite{mao2016refcocog} subset (3.0K) 
  & Single-target seg. \\
  & \cellcolor{gray!15} LISA++~\cite{yang2023lisa++} (2.0K); gRefCOCO~\cite{liu2023gres} (4.0K) 
  & \cellcolor{gray!15} Multi-instance seg. \\
\bottomrule
\end{tabular}%
}
\end{table} 
\label{app:methods_details}

\subsection{Offset-Aware Dataset Curation Details}
\label{app:offset_dataset}

\noindent \textbf{Region Definitions.}
Let \(\mathbf{M_\texttt{gt}}\in\{0,1\}^{H\times W}\) be the binary foreground mask. We place an \(n\times n\) grid and denote the pixel center of cell \((i,j)\) by \(\mathbf{c}_{i,j}=(x_{i,j},y_{i,j})^\top\).
To construct pools of candidate grid tokens, we employ morphology-based bands scaled according to the offset step size. Let \(\mathcal{K}_k\in\{0,1\}^{k\times k}\) represent a square structuring element with a side length of \(k\) pixels. We define:
\begin{equation}
\label{eq:kernels-dc}
\begin{aligned}
k_e &= \lfloor s_y\rfloor + 1, & \mathbf{E} &= \mathbf{M_\texttt{gt}} \ominus \mathcal{K}_{k_e}, \\
k_d &= 2\lfloor s_y\rfloor + 1, & \mathbf{D} &= \mathbf{M_\texttt{gt}} \oplus \mathcal{K}_{k_d},
\end{aligned}
\end{equation}
where \(\lfloor\cdot\rfloor\) denotes the floor operation, while \(\ominus\) and \(\oplus\) represent morphological erosion and dilation respectively.
A thin boundary band is additionally defined as:
\begin{equation}
\label{eq:boundary-dc}
\mathbf{B} = \operatorname{dilate}(\operatorname{grad}(\mathbf{M_\texttt{gt}}), \mathcal{K}_b),
\end{equation}
where \(\operatorname{grad}(\mathbf{M_\texttt{gt}})\) is the morphological gradient and \(b\) is a small width parameter.
By construction, \(\mathbf{E} \subset \mathbf{M_\texttt{gt}} \subset \mathbf{D}\): \(\mathbf{E}\) forms a step-sized interior buffer, \(\mathbf{D}\) creates a step-sized exterior halo, and \(\mathbf{B}\) captures edge uncertainty as a narrow boundary ribbon.

\vspace{1mm}
\noindent \textbf{Grid Point Categorization and Sampling.}
We define a one-step hit test to determine reachability:
\begin{equation}
\label{eq:hit-dc}
\operatorname{Hit}(i,j) \triangleq \exists\,\boldsymbol{\delta}\in\{-1,0,1\}^2: \mathbf{M_\texttt{gt}}(\mathbf{c}_{i,j} + \mathbf{S}\boldsymbol{\delta}) = 1.
\end{equation}
Each grid center is assigned to exactly one category via the hierarchical decision rule:
\begin{equation}
\label{eq:pool}
\begingroup
\setlength{\arraycolsep}{2pt} 
\operatorname{pool}(i,j)=
\begin{cases}
\text{Hard-Delete}, &
\begin{aligned}[t]
&\mathbf{B}(y_{i,j},x_{i,j})=1\\
&\mathbin{\wedge}\ \Mgt(y_{i,j},x_{i,j})=0\\
&\mathbin{\wedge}\ \neg\operatorname{Hit}(i,j),
\end{aligned}\\[0.2em]
\text{Inside}, & \mathbf{E}(y_{i,j},x_{i,j})=1,\\
\text{Ring},   &
\begin{aligned}[t]
&\mathbf{D}(y_{i,j},x_{i,j})=1\\
&\mathbin{\wedge}\ \Mgt(y_{i,j},x_{i,j})=0,
\end{aligned}\\
\text{Far},    & \text{otherwise.}
\end{cases}
\endgroup
\end{equation}

Following pool formation \(\mathcal{P}_{\mathrm{hard}} \to \mathcal{P}_{\mathrm{inside}} \to \mathcal{P}_{\mathrm{ring}} \to \mathcal{P}_{\mathrm{far}}\), we sample \(K \sim \pi_K\) grids per image with preferential selection from \(\mathcal{P}_{\mathrm{inside}}\) and \(\mathcal{P}_{\mathrm{ring}}\), while maintaining representation from all categories for robustness. Then, the complete construction process, detailed in Algorithm~\ref{alg:offset_dataset}, processes each image-mask-query triple to automatically produce conversational data containing grid tokens and their corresponding offset targets.
\begin{algorithm}[htbp]
\caption{Offset-Supervised Data Construction}
\label{alg:offset_dataset}
\small
\KwIn{Referring dataset $\mathcal{D}$; grid size $n$; offset granularity $m$; IoU threshold $\tau$}
\KwOut{JSONL conversations containing grids and offset targets}
\ForEach{$(I,\mathbf{M_\texttt{gt}},q)\in\mathcal{D}$}{
  Resize $I,\mathbf{M_\texttt{gt}}$ to $H\times W$; compute $s_x=W/m$, $s_y=H/m$, $\mathbf{S}=\operatorname{diag}(s_x,s_y)$\;
  \tcp{grid pools via morphology (cf.\ \eqref{eq:kernels-dc}--\eqref{eq:boundary-dc})}
  Compute $\mathbf{E},\mathbf{D},\mathbf{B}$; assign each grid cell $(i,j)$ to one of \textsc{Inside}/\textsc{Ring}/\textsc{Far}/\textsc{Hard-Delete} by rule \eqref{eq:pool}\;
  \tcp{Segmentation grids and offsets}
  Sample $K$ grids $\{(i_k,j_k)\}_{k=1}^K$ from the pools\;
  \For{$k=1$ \KwTo $K$}{
    Set $\mathbf{c}_k \gets \mathbf{c}_{i_k,j_k}$\;
    \uIf{$\mathbf{M_\texttt{gt}}(y_{i_k},x_{i_k})=1$}{emit \texttt{[OFF\_0\_0]}}
    \uElseIf{$\Hit(i_k,j_k)$}{
      pick $(\delta_u,\delta_v)\in\{-1,0,1\}^2$ with 
      $\mathbf{M_\texttt{gt}}(\mathbf{c}_k+\mathbf{S}\boldsymbol{\delta})=1$,
      and emit \texttt{[OFF\_\(\delta_u\)\_\(\delta_v\)]}
    }
    \Else{emit \texttt{<DELETE>}}
  }
  \tcp{Bounding-box corner offsets}
  Let $B^\star\!\gets\!\mathrm{BBox}(\mathbf{M_\texttt{gt}})$; jitter its TL/BR to grid corners $(i_{\mathrm{tl}},j_{\mathrm{tl}}),(i_{\mathrm{br}},j_{\mathrm{br}})$\;
  Evaluate all offset pairs for the two corners (apply $\mathbf{S}$-scaled displacements), obtain $\mathrm{IoU}_{\max}$\;
  \uIf{$\mathrm{IoU}_{\max}\ge\tau$}{emit the two corner offsets}
  \Else{emit \texttt{<DELETE>} for both corners}
  \tcp{Serialization}
  Write a JSONL sample with image tag, user prompt $q$ and grids (user turn), and the offsets (assistant turn)\;
}
\end{algorithm}

\subsection{Reward Details}
\label{app:gridtoken_reward}

\begin{figure*}[thbp]
    \centering
    \includegraphics[width=0.96\linewidth]{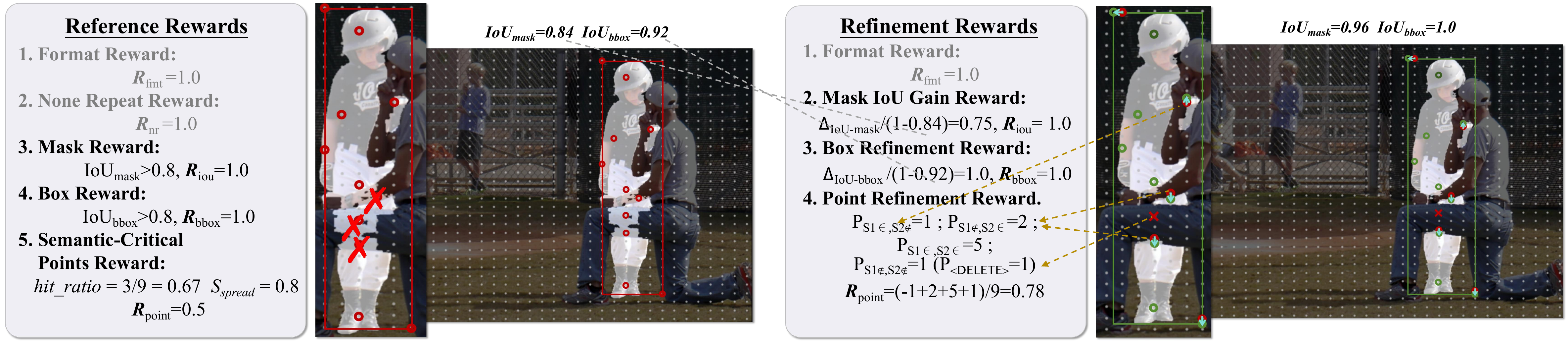}
    \caption{\textbf{Illustration of reward computation for grid token generation and refinement.} The diagram demonstrates how different reward components are calculated based on predicted outputs and ground-truth annotations.}
    \label{fig:reward_example}
\end{figure*}

\noindent \textbf{Multi-object Matching.} From each line in \texttt{<answer>}, we extract a predicted instance consisting of an optional box \(\hat{\vecb{b}}_p\in\mathbb{R}^{4}\) and a point set \(\mathcal{P}_p=\{\vecb{q}\}\subset\mathbb{R}^{2}\). Let there be \(P\) predictions and \(G\) ground-truth instances with binary masks \(\{\matb{M}_\texttt{gt}\}_{g=1}^{G}\) and tight boxes \(\{\vecb{b}_g\}_{g=1}^{G}\). We define pairwise similarities between prediction \(p\) and ground truth \(g\):

\noindent i) Box IoU:
\begin{equation}
\mathrm{IoU}_{p,g} \in [0,1] \label{eq:pair-iou}.
\end{equation}
\noindent ii) Point-hit ratio: the fraction of predicted points that land inside $\mathbf{M_\texttt{gt}}$,
\begin{equation}
H_{p,g} = \frac{1}{\max(1,|\mathcal{P}_p|)} \sum_{\mathbf{q}\in \mathcal{P}_p} \mathbbm{1}\{\mathbf{q}\in \mathbf{M_\texttt{gt}}\}\in[0,1].
\end{equation}
\noindent iii) Normalized $L_1$ box score:
\begin{equation}
S^{\ell_1}_{p,g} = \mathrm{clip}\!\left(1-\frac{\|\hat{\mathbf{b}}_p-\mathbf{b}_g\|_1/4}{\tau_{\ell_1}},\,0,\,1\right) \label{eq:pair-l1}.
\end{equation}
These are combined into a similarity used only for the assignment:
\begin{equation}
\mathrm{Sim}_{p,g} = \mathrm{IoU}_{p,g} + H_{p,g} +  S^{\ell_1}_{p,g},
\end{equation}
We solve a Hungarian assignment~\cite{kuhn1955hungarian} with costs $C_{p,g}=3-\mathrm{Sim}_{p,g}$, yielding matched pairs $\mathcal{M}\subseteq \{1..P\}\times\{1..G\}$.
We use $\tau_{\ell_1}{=}18\text{ px}$.

\vspace{1mm}
\noindent\textbf{Semantic-Critical Points Reward.}
For each $(p,g)\in\mathcal{M}$, we compute a key points quality:
\begin{equation}\label{eq:kq-pair}
F_{p,g} \;\triangleq\; S(m_p)\Bigl(w_H\,H_{p,g} \;+\; w_{\mathrm{spr}}\,\mathrm{Spread}_{p,g}\Bigr) \;-\; \lambda_m\,m_p.
\end{equation}
where \(H_{p,g}\) is the hit ratio, and \(\mathrm{Spread}_{p,g}\) rewards larger nearest-neighbor spacing normalized by object scale:

\begin{equation}
\label{eq:kq-spread}
\begin{aligned}
\bar d_p
&= \frac{1}{m_p}\sum_{i=1}^{m_p}\min_{j\neq i}\|\vecb{q}_i-\vecb{q}_j\|_2,\\[0.3em]
\mathrm{Spread}_{p,g}
&= \clip\!\big(\bar d_p / (\rho_s r_g),\,0,\,1\big).
\end{aligned}
\end{equation}
The multiplicative saturation \(S(m)=1-\exp(-m/m_0)\) discourages degenerate few-point outputs,
and the linear term \(\lambda_m m_p\) penalizes overly long point lists.
We aggregate across matches with point-count weighting:
\begin{equation}\label{eq:kq-agg}
T \;=\; \clip\!\left(
\frac{\sum_{(p,g)\in\mathcal{M}} m_p\,F_{p,g}}{\sum_{p=1}^{P}\max(1,m_p)}\!,\ 0,\,1\right).
\end{equation}
We set $w_H{=}0.6,\; w_{\mathrm{spr}}{=}0.4,\; \lambda_m{=}0.02,\; \rho_s{=}0.30$.

\vspace{1mm}
\noindent\textbf{Point Refinement Reward.}
Let $\mathbf{M}^{(k)}_{\mathrm{gt}}\subset\mathbb{Z}^2$ be the ground-truth mask of the $k$-th instance.
The coarse point set is $\mathcal{C}_{k}=\{\mathbf{c}_{k,p}\}_{p=1}^{P_k}$ and the refined set is
$\mathcal{C}^{\mathrm{off}}_{k}=\{\mathbf{c}^{\mathrm{off}}_{k,p}\}_{p=1}^{P_k}$,
with a one-to-one correspondence over $p$ (if a point is deleted, we keep its index $p$ and mark a delete flag). Define the inclusion indicators
$I_{k,p}=\mathbb{I}\!\big[\mathbf{c}_{k,p}\in \mathbf{M}^{(k)}_{\mathrm{gt}}\big]$,
$I^{\mathrm{off}}_{k,p}=\mathbb{I}\!\big[\mathbf{c}^{\mathrm{off}}_{k,p}\in \mathbf{M}^{(k)}_{\mathrm{gt}}\big]$.
The point-wise reward $s_{k,p}\in\{-1,0,1\}$ is
\begin{equation}
\begin{cases}
-1, & I_{k,p}=1 \land I^{\mathrm{off}}_{k,p}=0 \\[2pt]
+1, & I_{k,p}=0 \land I^{\mathrm{off}}_{k,p}=1 \\[2pt]
+1, & I_{k,p}=1 \land I^{\mathrm{off}}_{k,p}=1 \\[2pt]
+1, & I_{k,p}=0 \land  \text{\texttt{<DELETE>}} \wedge \mathcal{N}_{3\times3}(\mathbf{c}_{k,p}) \cap \mathbf{M_\texttt{gt}} = \emptyset\\[2pt]
0, & \text{otherwise.}
\end{cases}
\end{equation}
where $\mathcal{N}_{3\times3}(\mathbf{c}_{k,p})$ is the $3{\times}3$ neighborhood centered at $\mathbf{c}_{k,p}$.
The instance-level reward is obtained by averaging over all points of that instance. Fig.~\ref{fig:reward_example} provides a concrete example illustrating the reward computation process for better understanding.

\section{Additional Visualization Results}
\label{app:visualization_results}
\begin{figure*}[t]
    \centering
    \includegraphics[width=0.96\linewidth]{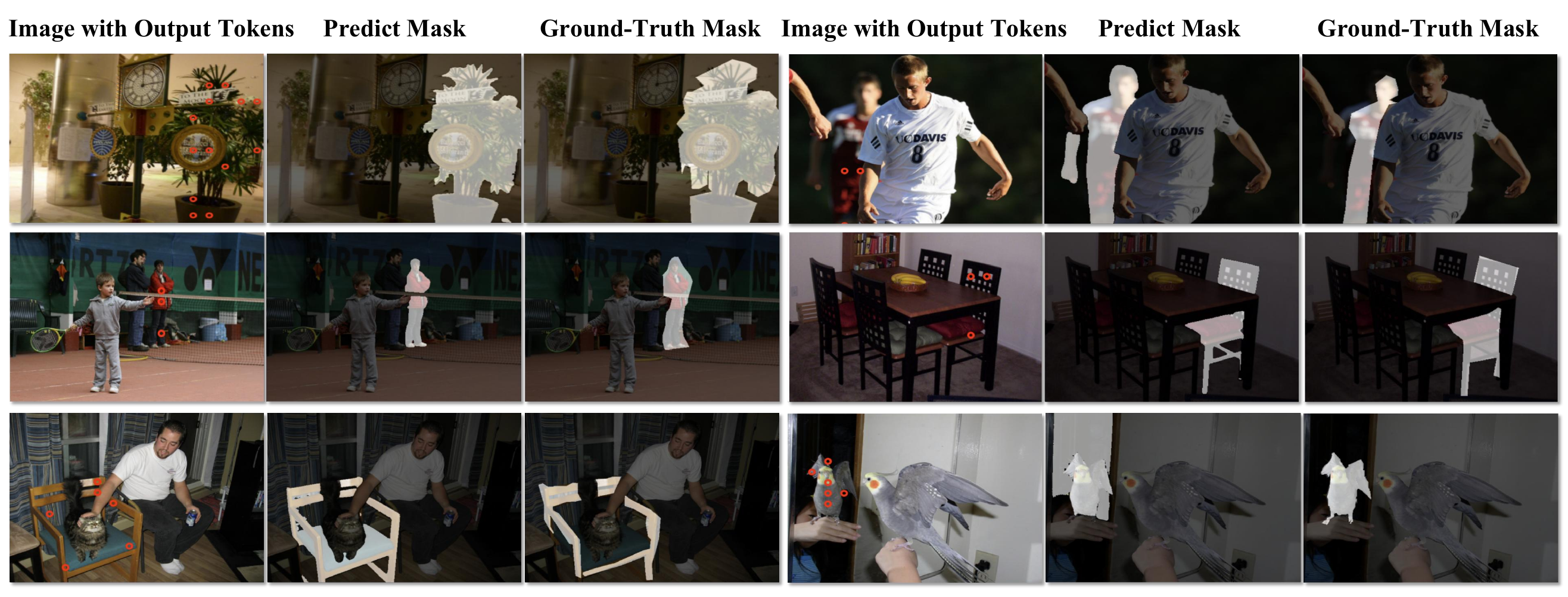}
    \caption{
    \textbf{More qualitative results of the segmentation task.} From top to bottom, the predictions are ordered by decreasing Intersection-over-Union (IoU) scores relative to the ground truth masks. }
    \label{fig:visual2}
    \vspace{20pt}
\end{figure*}

\begin{figure*}[htbp]
    \centering
    \includegraphics[width=0.96\linewidth]{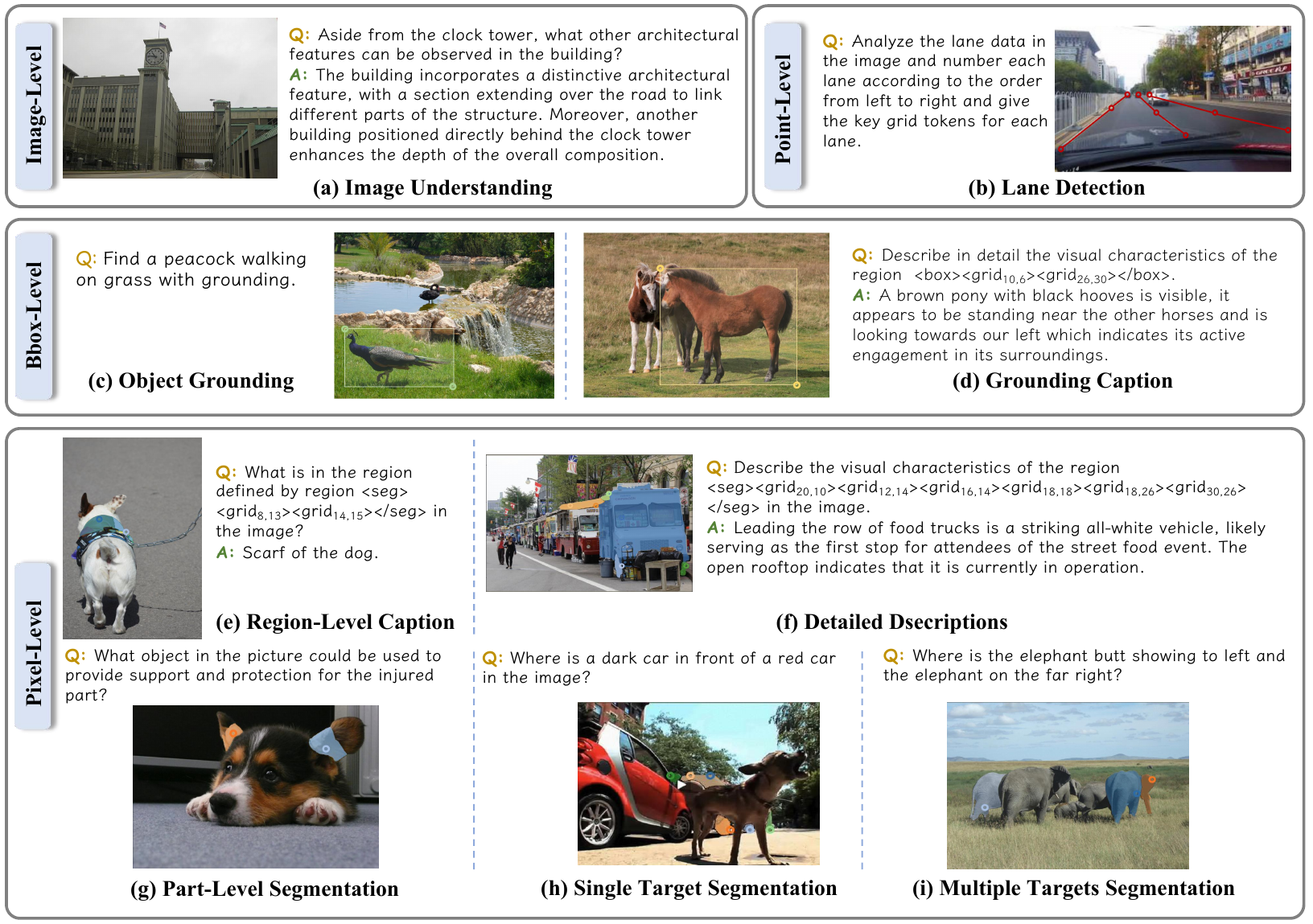}
    \caption{\textbf{Unified GETok representations across diverse vision-language tasks.} GETok provides a unified representation framework that handles diverse visual concepts without task-specific architectural modifications.
    }
    \label{fig:sft-benchmarks}
\end{figure*}
\noindent\textbf{Grid Tokens for Mask Representation.}
Fig.~\ref{fig:visual2} presents additional qualitative results comparing predicted grid tokens, output masks, and GT annotations. The results are organized from top to bottom, ranging from predictions that are more precise than the GT mask to some failure cases. These visualizations highlight the following key observations:

\noindent \emph{(1) High-Quality Predictions:} The model can generate highly accurate grid tokens, which align well with the GT masks. These results demonstrate the effectiveness of grid tokens in precisely localizing and referring to objects in complex scenes.

\noindent(2) \emph{Failure Cases:} In some cases, accurate grid-token predictions still yield imperfect masks due to discrepancies in SAM’s mask decoding. Nonetheless, as discussed in Sec.~\ref{sup:mask}, this training-free decoding remains advantageous compared to training task-specific mask decoders. Introducing offset tokens further mitigates these errors by refining point locations and aligning the decoded masks more closely with object boundaries.

The qualitative results underscore the robustness of grid tokens as a referring representation, even in cases where segmentation performance is suboptimal.

\vspace{1mm}
\noindent\textbf{SFT Benchmarks Qualitative Results.}
Fig.~\ref{fig:sft-benchmarks} demonstrates the unified representation capability of GETok across diverse vision-language tasks. Our approach establishes a cohesive framework that processes various query types through a consistent token vocabulary, spanning image-, point-, box-, and mask-level formats while eliminating the need for task-specific output heads.

\begin{figure*}[thbp]
    \centering
    \includegraphics[width=0.9\linewidth]{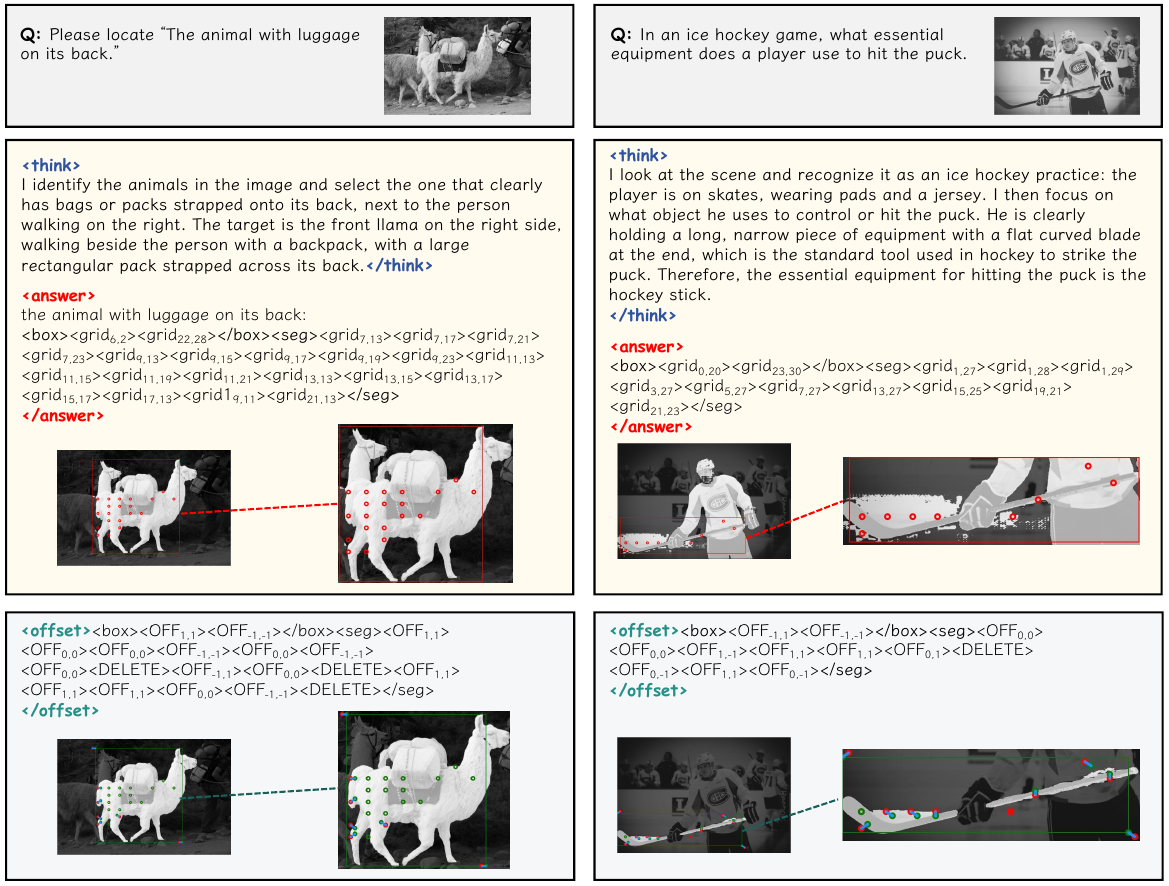}
    \caption{\textbf{More qualitative results of the self-improving mechanism.} Additional examples demonstrate how GETok establishes initial spatial proposals through grid tokens (red dots) and enables fine-grained adjustments via offset tokens (blue arrows), showing effective handling of objects across scales with enhanced precision on small targets.}
    \label{fig:case2}
\end{figure*}

\vspace{1mm}
\noindent\textbf{Self-Improving Mechanism.}
Fig.~\ref{fig:case2} presents additional qualitative examples demonstrating the propose-and-refine workflow of GETok for fine-grained mask prediction. The left panel shows that for interior points unambiguously inside the mask, the model correctly maintains their positions without unnecessary adjustments, focusing refinement efforts exclusively on boundary regions. The right panel illustrates a failure case primarily caused by erroneous refinement decisions resulting from initial tokens placed near misleading edge features. These examples collectively highlight the method's capacity to maintain accurate localization through coordinated grid and offset token operations, even in challenging scenarios.
\vspace{8em}
\label{app:case}


\end{document}